\RequirePackage{fix-cm}
\documentclass[twocolumn]{svjour3}          
\smartqed  
\usepackage{graphicx}
 \journalname{Neural Computing and Applications}

\usepackage{rotating}
\usepackage{makecell}
\usepackage{array}
\usepackage{eqnarray,amsmath}
\usepackage{amsmath}
\usepackage{graphicx}
\usepackage{esvect}
\usepackage{tabularx}
\usepackage{longtable}
\usepackage{lscape} 
\usepackage{amssymb}
\usepackage{amsmath,url,epsfig,color,booktabs,multirow}
\usepackage[table]{xcolor}

\begin{document}

\title{Learning Visual Representations with Optimum-Path Forest and its Applications to Barrett's Esophagus and Adenocarcinoma Diagnosis
}

\titlerunning{OPF Visual Representations for BE and Adenocarcinoma Diagnosis}        

\author{Luis~A.~de~Souza~Jr.         \and
        Luis~C.~S.~Afonso           \and 
        Alanna~Ebigbo               \and
        Andreas~Probst              \and
        Helmut~Messmann           \and
	Robert~Mendel               \and
        Christian~Hook              \and    
        Christoph~Palm              \and
        Jo\~ao~P.~Papa              
}


\institute{Luis~A.~de~Souza~Jr. and Luis~C.~S.~Afonso   \at
              Department of Computing, Federal University of S\~ao  Carlos - UFScar, Brazil \\
              \email{\{luis.souza,luis.afonso\}@dc.ufscar.br}           
           \and
	   Alanna~Ebigbo, Andreas~Probst and Helmut~Messmann \at 
                Medizinische Klinik III, Klinikum Augsburg - Germany\\
                \email{\{alanna.ebigbo,andreas.probst,helmut.messmann\}@-\\klinikum-augsburg.de}
                \and
           Robert~Mendel, Christian~Hook and Christoph~Palm \at
              Regensburg Medical Image Computing (ReMIC), Ostbayerische Technische Hochschule Regensburg - OTH Regensburg, Germany \\
              \email{robert.mendel@st.oth-regensburg.de,\\\{christian.hook,christoph.palm\}@oth-regensburg.de}
           \and
           Jo\~ao~P.~Papa \at 
                Department of Computing, S\~ao Paulo State University - UNESP, Brazil\\
                Phone/fax: +55-14-3103-6079\\ 
                \email{papa@fc.unesp.br}
}

\date{Received: date / Accepted: date}

\maketitle

\begin{abstract}
Considering the rose number of the Barret's esophagus (BE) number in the last decade, and its expectation of continue increasing, methods that can provide an early diagnosis of dysplasia in BE diagnosed patients may provide a high probability of cancer remission. The limitations related to traditional methods of BE detection and management encourage the creation of computer-aided tools to assist in this problem. In this work, we introduce the unsupervised Optimum-Path Forest (OPF) classifier for learning visual dictionaries in the context of Barrett's esophagus (BE) and automatic adenocarcinoma diagnosis. The proposed approach was validated in two datasets (MICCAI 2015 and Augsburg) using three different feature extractors (SIFT, SURF, and \textcolor{black}{the not yet applied to the BE context A-KAZE}), as well as five supervised classifiers, including two variants of the OPF, Support Vector Machines with Radial Basis Function and Linear kernels, and a Bayesian classifier. Concerning MICCAI 2015 dataset, the best results were obtained using unsupervised OPF for dictionary generation using supervised OPF for classification purposes and using SURF feature extractor with accuracy nearly to $78\%$ for distinguishing BE patients from adenocarcinoma ones. Regarding the Augsburg dataset, the most accurate results were also obtained using both OPF classifiers but with A-KAZE as the feature extractor with accuracy close to $73\%$. The combination of feature extraction and bag-of-visual-words techniques showed results that outperformed others obtained recently in the literature, as well as we highlight new advances in the related research area. \textcolor{black}{ Reinforcing the significance of this work, to the best of our knowledge, this is the first one that aimed at addressing computer-aided BE identification using bag-of-visual-words and OPF classifiers, being this application of unsupervised technique in the BE feature calculation the major contribution of this work. It is also proposed a new BE and adenocarcinoma description using the A-KAZE features, not yet applied in the literature.}
\keywords{Barrett's esophagus \and optimum-path forest \and machine learning \and adenocarcinoma \and image processing}
\end{abstract}


\section{Introduction}
\label{s.introduction}

Pattern classification has been paramount in the last decades, mainly due to the increasing number of applications that require some intelligent-decision-making mechanism. The standard pipeline adopted for so many years follows a robust but straightforward workflow: (i) feature extraction, (ii) model learning, and (iii) classification outcomes. The former step can be performed using handcrafted features or information learned through deep learning approaches. In this latter case, one may not know what kind of information the model is learning, since the set of outcome values that minimizes some loss function is the one employed in the model learning step. Handcrafted features require a more knowledgeable personnel, which is usually in charge of selecting and extracting features that matter when performing pattern classification.

\begin{sloppypar}Describing images using their most important information, the so-called ``points of interest" (PoIs) or key points, has been an active area of interest by many researchers worldwide. Notable approaches have been proposed in the literature to compute those points, which somehow aim at capturing subtle information that is less variant to geometric transformations such as rotation, and translation, among others. Scale-Invariant Feature Transform (SIFT)~\cite{Lowe:04}, Speeded-Up-Robust-Features (SURF)~\cite{Bay:08}, and Accelerated-KAZE features (A-KAZE)~\cite{AlcantarillaBMVC:13} are some examples.\end{sloppypar}

\begin{sloppypar}However, the main problem related to the mentioned approaches concern the final feature vector. Since the number of PoIs may vary from one image to another, the feature vectors used to represent the images shall have different dimensions. To overcome this issue, an additional step called ``quantization"\ is required (some works refer to this step as the ``codebook generation"~\cite{FeiFeiCVPR:05}). In a nutshell, given a training set composed of PoIs extracted from all training images, we can build a ``bag" (i.e., a visual dictionary) with the most representative PoIs (from now on called ``visual words"). Further, for each training image and each of its PoIs, we can find the ``closest"\ visual word in the bag and build up a histogram that stores the number of times a visual word is nearest to each PoI from the training images. Therefore, the final feature vector of each training image will be that histogram with a dimensionality that corresponds to the number of visual words (i.e., the size of the dictionary or bag). Essentially, that is the main reason such approaches are usually referred to as ``bag-of-visual-words" (BoVW)~\cite{CsurkaECCV:04}.\end{sloppypar}

Bag-of-visual-words have been widely used in the literature for a number of different purposes, such as video-based action recognition~\cite{PengCVIU:16},  retinal health diagnosis~\cite{KohCBM18}, and perivascular spaces categorization in brain data~\cite{GonzalezICIAR:16}, among others. Nonetheless, one still has two problems to face regarding the BoVW approach: (i) how to find out the most representative visual words, and (ii) how to establish a proper bag size, i.e., the number of visual words. Notice that both issues are pretty much crucial since they are in charge of the feature vector composition and dimensionality.

To cope with the first issue, i.e., finding out the most representative visual words, two approaches are commonly used: (i) random sampling and (ii) clustering. The former randomly selects a given number of visual words to compose the bag. On the other hand, clustering-based approaches make use of some unsupervised learning algorithm  (usually k-means) to group the visual words, and the most representative ones (i.e., centroids) are elected to compose the dictionary~\cite{AfonsoICIP:12}. However, randomly choosing visual words does not lead to good results, and the usage of certain unsupervised learning algorithms turns out to be a problem since most of them require the number of clusters (i.e., the bag size) beforehand.

Therefore, clustering techniques that do not require a priori information about the data are usually preferred. Among several techniques that have been proposed in the literature, one is gaining attention daily due to its effectiveness and efficiency in different research areas. The Optimum-Path Forest (OPF) is a framework for the design of pattern classifiers based on graph partition. In short, OPF-based classifiers work on a reward-competition process, in which previously selected samples called ``prototypes"\ try to conquer other samples by offering them optimum-path costs. Once a sample is conquered by another one, it receives its label and a ``mark" (i.e., a predecessor map) that reveals its conqueror. The Optimum-Path Forest framework comprises supervised~\cite{PapaIJIST:09,PapaPR:12,PapaPRL:17}, unsupervised~\cite{RochaIJIST:09}, and semi-supervised~\cite{AmorimPR:16} versions that have been widely employed in a number of applications, from remote sensing~\cite{PisaniTGRS:14,NakamuraIEEETGRS:14} to human intestinal parasites identification~\cite{SuzukiTBE:13}, just to cite a few.

One particular strength of unsupervised OPF concerns the fact it does not require the number of clusters beforehand, i.e., it finds clusters on-the-fly. Such feature is quite interesting in the context of BoVW generation since we skip the problem of choosing suitable bag sizes. As far as we are concerned, only two works attempted at using OPF in the context of BoVW: (i) Papa and Rocha~\cite{PapaICIP:11} evaluated the supervised OPF for image categorization using visual words, and further (ii) Afonso et al.~\cite{AfonsoICIP:12} studied the impact of using unsupervised OPF for learning proper visual dictionaries.

We are particularly interested \textcolor{black}{in the application of such technique} for the recognition of Barrett's esophagus (BE), which happens to be a side effect of some reflux diseases. \textcolor{black}{BE comprises a very severe and growing disease in the last decades, and} since BE is often not identified properly at the early stages, it may evolve to a more aggressive version, and even to cancer. \textcolor{black}{However, the early diagnosis of dysplastic tissue in BE diagnosed patients may provide very high rates of remission after the treatment~\cite{Dent2011,Sharma2016,Phoa555}. There are several endoscopic techniques for the BE diagnosis and detection, such as chromoendoscopy and narrow-band imaging, but the human screening for the injured region definition is still often misclassified by endoscopists, once the region does not present enough goblet cells in biopsy or the experts refuse to use the recommended procedure for extensive biopsies~\cite{Sharma20152209}. Moreover, computer-assisted diagnosis may bring precision and accuracy to the BE screening and evaluation, once this task can be very influenced by the human factor~\cite{Souza_SIBGRAPI:17,SouzaJr2018203,Souza:BVM2017}. } To the best of our knowledge, only one very recent work coped with BE identification using OPF. Souza et al.~\cite{Souza_SIBGRAPI:17} introduced the supervised OPF for Barrett's esophagus automatic identification using features based on BoVW. The authors considered both random- and $k$-means-based sampling strategies to build the visual dictionaries and then used OPF for classification purposes. 

Some works that dealt with endoscopic image analysis can be referred to as well, but that is still an emerging area of research~\cite{SouzaJr2018203}. Seibel et al.~\cite{SeibelTBE:08} developed a low-cost but high-performance technology to assist the diagnosis of BE and esophageal cancer. However, their primary contributions rely on hardware advances rather than software ones. The work presented by van der Sommen~\cite{SommenEndoscopy:16} aimed at using machine learning techniques to detect early neoplasia in Barrett's esophagus, and Swager et al.~\cite{SwagerGastroenterology:16} addressed the very same context mentioned above but using volumetric laser endoscopy images. 

\begin{sloppypar}Klomp et al.~\cite{KlompSPIE:17} proposed new features for computer-aided Barrett's esophagus identification, and Hassan and Haque~\cite{HassanCBM:15} used endoscopy videos obtained from wireless capsules to assess gastrointestinal hemorrhages. Later on, Segu\'i et al.~\cite{Seguia2016} used the same source of images (i.e., wireless capsules) together with Deep Convolutional Neural Networks for intestine motility characterization. Mendel et al.~\cite{Mendel2017} started the study of deep learning application to the BE and adenocarcinoma evaluation problem.\end{sloppypar}

\textcolor{black}{The major problems around the computer-assisted systems developed for the BE and adenocarcinoma evaluation are related to the type of technique to provide a correct description of the injured areas and which classification techniques should be designed for the problem. These problems are related to all proposed works, and considering the high potential in this research area, new ways to describe the injured areas (that are very similar), and the evaluation of different classifiers can deliver important and substantial improvements to the precision and correct differentiation of both.}

\textcolor{black}{As one can observe, Barrett's esophagus automatic identification using machine learning techniques presents a growing interest in the last years. Therefore, there is plenty of room for new works that employ techniques that were not considered in such a context. In this work, we extended and outperformed the approach proposed by Souza et al.~\cite{Souza_SIBGRAPI:17} by learning proper visual dictionaries using unsupervised OPF, as well as we introduced a variant of supervised OPF (OPF$_{knn}$) proposed by Papa et al.~\cite{PapaPRL:17} in the context of BE identification. The OPF was never applied to such problem in the visual learning step, and this could deliver, besides the novelty in the feature vector calculation, a new way to evaluate the key points provided by the feature extraction techniques. Last but not least, we introduce the A-KAZE feature extraction technique for the calculation of the key points, for comparison with SURF and SIFT, previously adopted for the BE and adenocarcinoma differentiation context~\cite{SouzaJr2018203}.} The results presented in this paper are close to some state-of-the-art recognition rates~\cite{Mendel2017}, and it features recent advances to BE automatic identification by means of machine learning and computer vision.Therefore, the main contributions of this paper are five-fold:
\begin{itemize}
    \item to extend and outperform the recent results obtained by Souza et al.~\cite{Souza_SIBGRAPI:17} \textcolor{black}{in which the evaluation of BE and adenocarcinoma context were performed using: (i) SURF and SIFT techniques for key points calculation, (ii) $k$-means and random techniques for the bag-of-visual-words calculation, and (iii) OPF and SVM classifiers for the classification task};
    \item to introduce OPF$_{knn}$~\cite{PapaPRL:17} for BE and adenocarcinoma automatic diagnosis, \textcolor{black}{considering that Souza et al.~\cite{Souza_SIBGRAPI:17} employed only the complete graph version of OPF classifier for the classification task};
    \item to introduce A-KAZE features for the aforementioned context, \textcolor{black}{once such technique has been largely applied in the literature for image description and retrieving}; 
    \item to extend the work by Afonso et al.~\cite{AfonsoICIP:12} with a more robust evaluation of the unsupervised OPF for learning visual dictionaries;
    \item \textcolor{black}{to introduce a new representation of feature extraction techniques (such as SURF and SIFT) based on their most representative words in the feature space using the OPF clustering technique.}
\end{itemize}

The remainder of this paper is organized as follows. Sections~\ref{s.opf} to~\ref{s.methodology} present a theoretical background of unsupervised OPF and the methodology adopted in this work, respectively. Section~\ref{s.experiments} discusses the experiments, and Section~\ref{s.conclusions} states conclusions and future works.



\section{Unsupervised Learning with Optimum-Path Forest}
\label{s.opf}

In this section, we briefly present the theoretical background related to unsupervised OPF, which is used to learn proper visual dictionaries.

Let ${\cal D}=\{\textbf{x}_1,\textbf{x}_2,\ldots,\textbf{x}_m\}$ be an unlabeled dataset such that $\textbf{x}_i\in\Re^n$ stands for a feature vector extracted from some sample (i.e., images in our case) related to the problem to be addressed. Additionally, let ${\cal G}=({\cal D},{\cal A}_k)$ be a graph derived from that dataset, which means ${\cal D}$ denotes the set of graph nodes (i.e., vertices) and ${\cal A}_k$ stands for a $k$-nearest neighbors adjacency relation.

In a nutshell, the OPF working mechanism is based on a reward-competition problem, where some samples called ``prototypes"\ \textcolor{black}{employ} a competitive process among themselves to conquer the other samples from the dataset ${\cal D}$. Such competition ends up partitioning ${\cal D}$ into optimum-path trees (OPTs), which are rooted at each prototype node. It is worth mentioning that a sample that belongs to a given OPT is more ``strongly connected"\ to the root and samples of that tree than to any other in the forest (i.e., a collection of all trees in the graph).

At a glance, the whole process can be summarized in the following steps:

\begin{enumerate}
\item To establish a proper neighborhood size and build up ${\cal A}_k$ (i.e., to find out ``suitable"\ $k$ values);
\item To elect the prototypes and Learning Visual Representations with Optimum-Path Forest and its Applications to Barrett's esophagus and Adenocarcinoma Diagnosis
\item To start the competition process.
\end{enumerate}

Concerning step 1), a number of different approaches to cope with the task could be considered. Rocha et al.~\cite{RochaIJIST:09} proposed to compute the best value of $k$ (i.e., the neighborhood size), say that $k^\ast$, as the one that minimizes the normalized graph cut, which is a measure that considers both the dissimilarity between clusters as well as the similarity within the groups of samples~\cite{ShiTPAMI:00}.

Soon after computing $k^\ast$, the next move concerns finding the prototypes (i.e., step 2), also known as the ``roots of the trees". Such essential samples are in charge of ruling the competition process that ends up partitioning the graph into OPTs (i.e., clusters). Those samples will be used as the visual words to compose the final dictionary, as further discussed.

The supervised OPF proposed by Papa et al.~\cite{PapaIJIST:09} elects the prototypes as the nearest samples from different classes, which can be accomplished by computing a Minimum Spanning Tree (MST) over the training graph. Then, the samples from different classes that are connected in the MST are marked as prototypes. However, unsupervised OPF does not make use of labeled datasets, which motivated Rocha et al.~\cite{RochaIJIST:09} to elect the prototypes as the samples that are located at the center of the clusters. Such samples can be computed by assigning a density score $\rho(\textbf{x}_i)$ for each dataset sample $\textbf{x}_i\in{\cal D}$. That score is computed using a probability density function (pdf) given by a Gaussian distribution considered in the neighborhood of each sample as follows:
\begin{equation}
	\label{e.density}
	\rho(\textbf{x}_i)=\frac{1}{\sqrt{2\pi\sigma^2}k}\sum_{\forall \textbf{x}_j\in {\cal A}_k(\textbf{x}_i)}\exp\left(\frac{-d(\textbf{x}_i,\textbf{x}_j)}{2\sigma^2}\right),
\end{equation}
where $i\neq j$ and $\sigma=d_{max}/3$. In this case, $d_{max}$ stands for the maximum arc-weight in $G$. Using such formulation, $\rho(\textbf{x}_i)$ considers all adjacent nodes for the probability computation purposes since a Gaussian function covers $99.7\%$ of the samples within $d(\textbf{x}_i,\textbf{x}_j)\in[0,3\sigma]$. 

After computing Equation~\ref{e.density} for all nodes, the competition process among samples can take place. Each density value will be used to populate a priority queue, where the idea of the unsupervised OPF algorithm is to end up maximizing the cost of each sample, and thus partitioning the graph. 

The definition of ``cost"\ is based on paths on graphs, i.e., a sequence of adjacent samples with no cycles. Let $\pi_{\textbf{x}_i}$ be a path with terminus at sample $\textbf{x}_i$ and starting from some root ${\cal R}(\textbf{x}_i)$, where ${\cal R}$ stands for the set of prototype samples. Additionally, let $\pi_{\textbf{x}_i}=\langle \textbf{x}_i\rangle$ be a trivial path (i.e., a path composed of a single sample) and $\pi_{\textbf{x}_i}\cdot \langle
\textbf{x}_i,\textbf{x}_j\rangle$ the concatenation of $\pi_{\textbf{x}_i}$ and the arc $(\textbf{x}_i,\textbf{x}_j)$ such that $i\neq j$.

The OPF algorithm assigns to each path $\pi_{\textbf{x}_i}$ a value $f(\pi_{\textbf{x}_i})$ given by a connectivity function $f:{\cal X}\rightarrow \Re$. In this context, a path $\pi_{\textbf{x}_i}$ is considered optimum if $f(\pi_{\textbf{x}_i})\geq f(\tau_{\textbf{x}_i})$ for any other path $\tau_{\textbf{x}_i}$. Such sort of functions are known as ``smooth functions", and they figure important constraints that ensure the theoretic correctness of the OPF algorithm~\cite{FalcaoIEEEPAMI:04}.
 
Among different path-cost functions that have been proposed in the literature, unsupervised OPF employs the following formulation for $\forall\textbf{x}_i,\textbf{x}_j\in{\cal D}$ such that $i\neq j$:
\begin{eqnarray}
f(\langle \textbf{x}_i \rangle) & = & \left\{ \begin{array}{ll} 
    \rho(\textbf{x}_i)           & \mbox{if $\textbf{x}_i \in {\cal R}$} \\
    \rho(\textbf{x}_i) - \delta  & \mbox{otherwise,}
 \end{array}\right.
\label{e.path_cost_function1}
\end{eqnarray}
and
\begin{equation}
f(\pi_{\textbf{x}_i}\cdot \langle\textbf{x}_i,\textbf{x}_j\rangle) = \min \{f(\pi_{\textbf{x}_i}), \rho(\textbf{x}_j)\},
\label{e.path_cost_function2}
\end{equation}
where $\delta = \min_{\forall (\textbf{x}_i,\textbf{x}_j)\in {\cal A}_k | \rho(t) \neq \rho(s) } |\rho(t)-\rho(s)|$. 
In a nutshell, $\delta$ stands for the smallest quantity required to avoid plateaus in the regions nearby the prototypes (i.e., areas with the highest density).  

Among all possible paths $\pi_{\textbf{x}_i}$ from the maxima of the pdf, the method assigns to sample $\textbf{x}_i$ a final path whose minimum density value along it is maximum. Such final path value is represented by a cost map ${\cal C}$, as follows: 
\begin{equation}
\label{e.cost_map}
	{\cal C}(\textbf{x}_i)=\max_{\forall \pi_{\textbf{x}_j}\in ({\cal D},{\cal A}_k), i\neq j} \{f(\pi_{\textbf{x}_j}\cdot \langle\textbf{x}_j,\textbf{x}_i\rangle)\}.
\end{equation}

The OPF algorithm maximizes the connectivity map ${\cal C}(\textbf{x}_i)$, $\forall \textbf{x}_i\in{\cal D}$, by computing an optimum-path forest over the dataset. Such forest is encoded as a predecessor map ${\cal P}$ with no cycles that assigns to each sample $\textbf{x}_i\notin {\cal R}$ its predecessor ${\cal P}(\textbf{x}_i)$ in the optimum path from ${\cal R}$, or a marker $nil$ when $\textbf{x}_i\in {\cal R}$.

\begin{figure}[!htb]
\begin{center}
 \begin{tabular}{c}
  \includegraphics[scale=0.55]{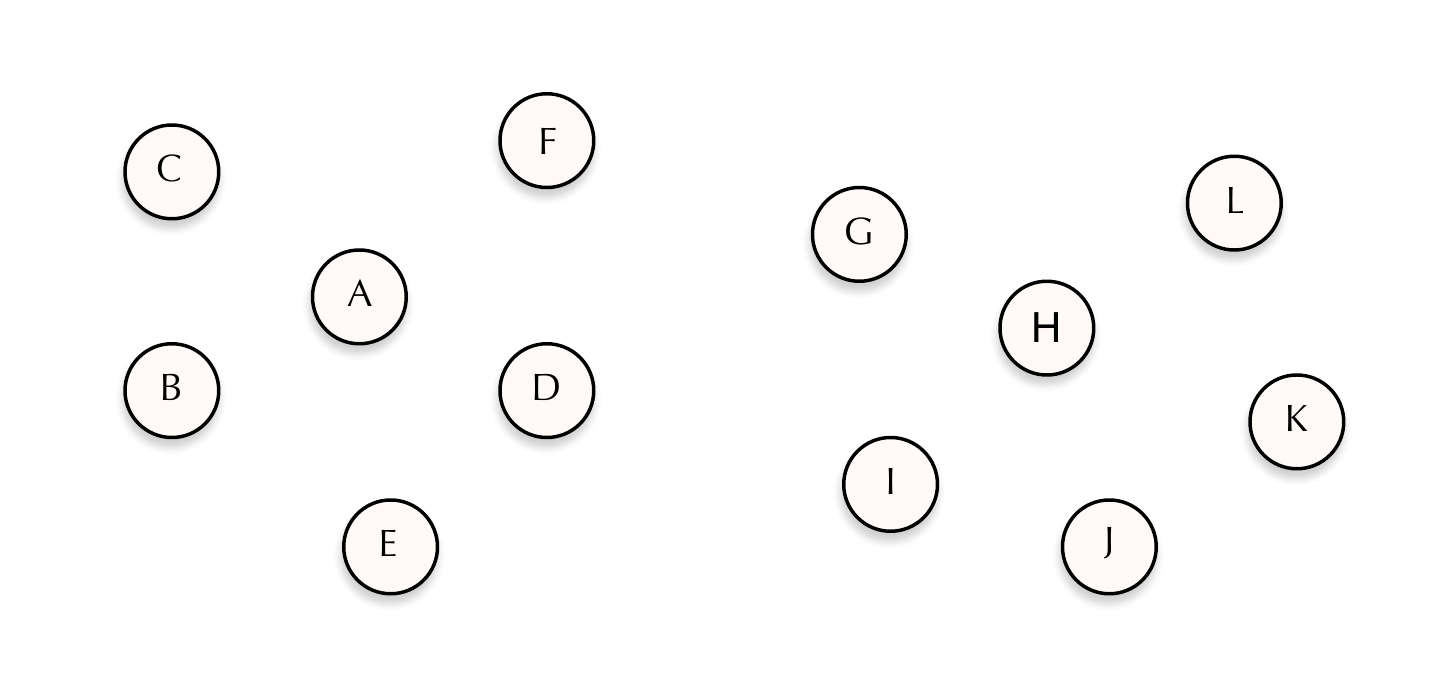} \\
  (a)
 \end{tabular}
 \begin{tabular}{c}	
 \includegraphics[scale=0.55]{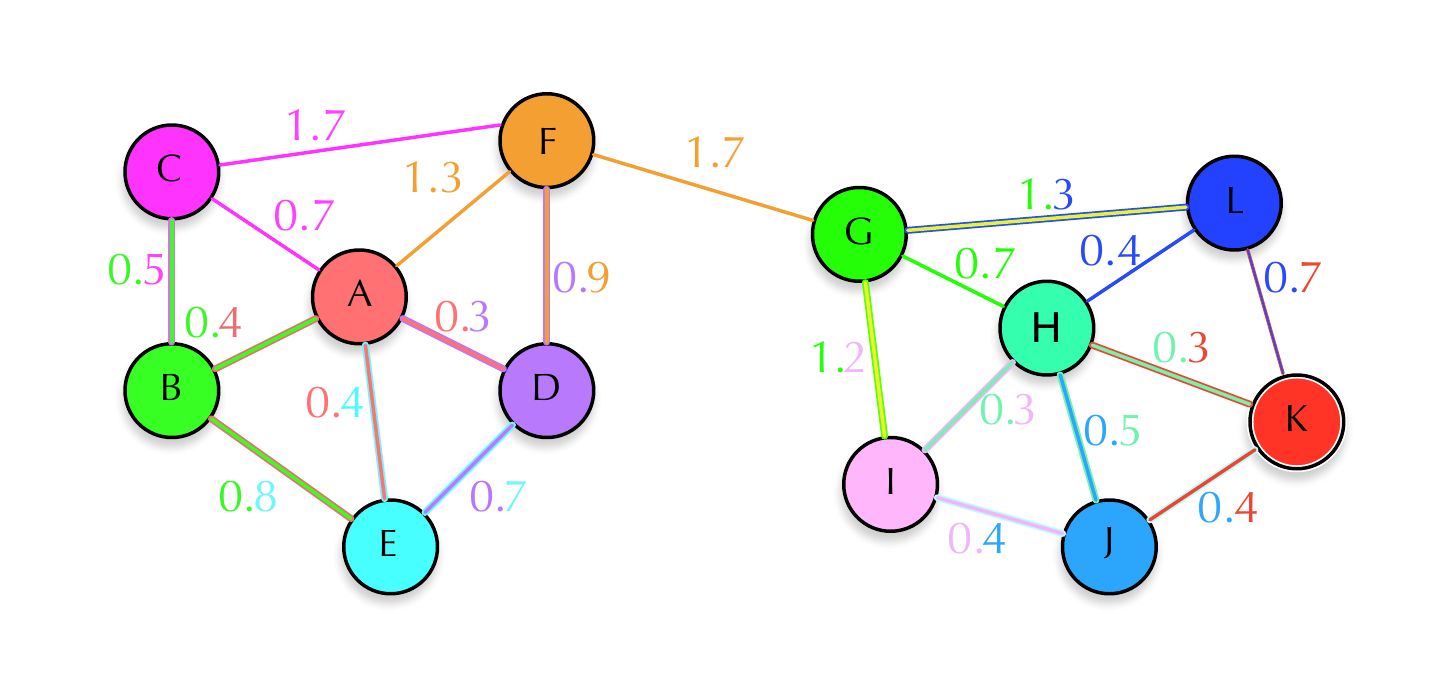}\\
 (b)
 \end{tabular}
\end{center}
 \caption{Toy example: (a) unlabeled dataset and its (b) $3$-nearest neighbors graph.} 
 \label{f.image_1}
\end{figure}

\begin{figure}[!htb]
\begin{center}
 \begin{tabular}{c}
  \includegraphics[scale=0.55]{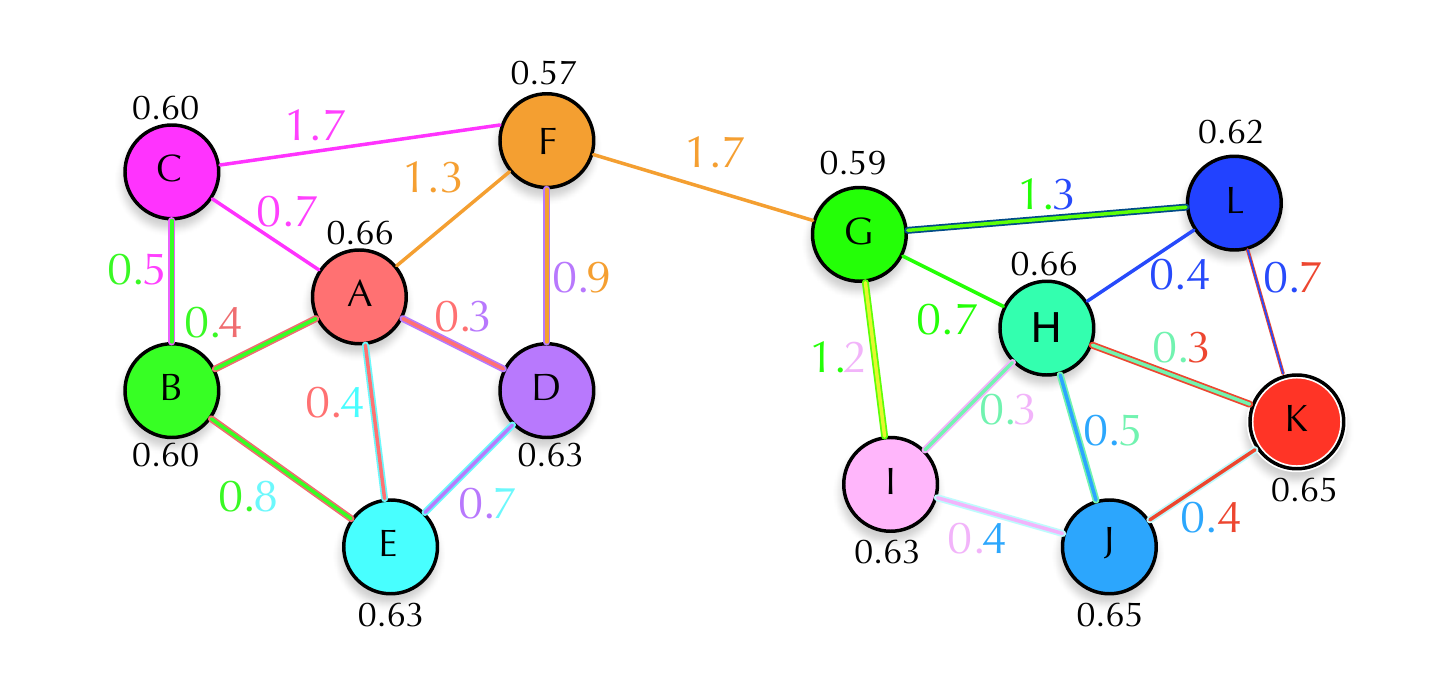} \\
 \end{tabular}
\end{center}
 \caption{Computing the densities of each graph node according to its $3$-neighborhood. The values under/over the nodes stand for their density values computed using Equation~\ref{e.density}.} 
 \label{f.image_2}
\end{figure}

\begin{figure}[!htb]
\begin{center}
 \begin{tabular}{c}
  \includegraphics[scale=0.55]{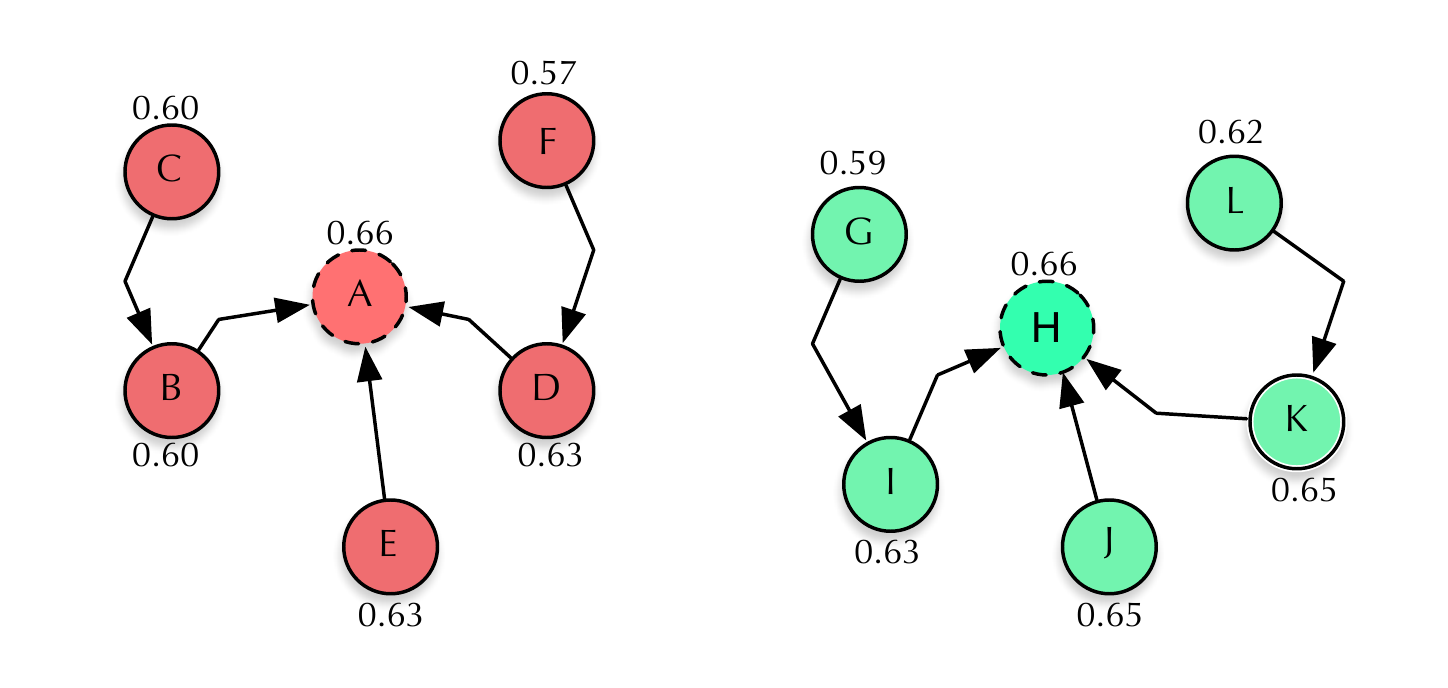} \\
 \end{tabular}
\end{center}
 \caption{Resulting optimum-path forest with two clusters and prototypes highlighted.} 
 \label{f.image_3}
\end{figure}

\begin{sloppypar}Figures~\ref{f.image_1} to~\ref{f.image_3} depict a toy example concerning the unsupervised OPF working mechanism. Figures~\ref{f.image_1}a and~\ref{f.image_1}b illustrate an unlabeled dataset and its $3$-nearest neighbors graph, respectively (we assume $k=3$ to explain step 1). For the sake of visualization purposes, we assigned the same color to each graph node and the arcs corresponding to its $3$-nearest neighbors.\end{sloppypar}

Notice the arcs are also weighted by the distance (e.g., Euclidean distance) among their corresponding nodes. One can observe that some arcs and their weights are double-colored, which means their corresponding nodes share the very same $3$-neighborhood.

Figure~\ref{f.image_2} illustrates the density computation step to further elect the prototypes (i.e., step 2). Therefore, given the arc-weights depicted in Figure~\ref{f.image_1}b, we can use Equation~\ref{e.density} to compute $\rho(\textbf{x}_i)$, $\forall \textbf{x}_i\in{\cal D}$. Notice the density values are computed over the adjacency relation encoded by ${\cal A}_k$. One can realize that the samples located at the center of the clusters tend to be the ones with the highest value of $\rho$ since they are connected by smaller arc-weights.

The density values are then stored in a priority queue (i.e., a max-heap) that pops out the sample $\textbf{x}_i$ with the highest $\rho(\textbf{x}_i)$. Concerning the toy example depicted in Figure~\ref{f.image_2}, the first sample to come out of the queue is either `H'\ or `A'\ since both have the highest densities. Suppose `H'\ has been added first to the queue. Since it has no predecessor, it is added to the set ${\cal R}$ and assigned $f(H) = \rho($H$) = 0.66$ according to Equation~\ref{e.path_cost_function1}.

Further, the competition process (i.e., step 3) takes place. In short, sample `H'\ evaluates its neighbors `I', `J', and `K'\ to offer better costs to them (i.e., costs that are greater than the ones they have already). Therefore, one has $f($H$\cdot\langle $H$,$I$\rangle)=\setminus min\{0.66,0.63\} = 0.63$, $f($H$\cdot\langle $H$,$J$\rangle)=\setminus min\{0.66,0.65\} = 0.65$, and $f($H$\cdot\langle $H$,$K$\rangle)= \setminus min\{0.66,0.65\} = 0.65$. Since the costs offered by `H'\ are greater or equal than the costs of its neighbors, they are conquered by sample `H'. Such process is encoded by the aforementioned predecessor map ${\cal P}$, i.e., after this first move of sample `H', one has that ${\cal P}($I$)=$ H, ${\cal P}($J$)=$ H, and ${\cal P}($K$)=$ H. 

The next sample to start the competition process is sample `A', and the very same process mentioned earlier is repeated until all samples have played in the competition process. The resulting optimum-path forest is depicted in Figure~\ref{f.image_3}. Notice one can obtain a different number of clusters based on the value of $k_{max}$. In this toy example, we obtained two clusters, which are labeled with the same color of its prototype/root of the tree (i.e., the dashed nodes `A'\ and `H').

The unsupervised OPF algorithm finds the number of the clusters on-the-fly, which means there is no need to have such information beforehand. The only parameter that needs to be set is the $k_{max}$, which constraints the search for suitable neighborhood sizes. One can observe that the knowledge required to set $k_{max}$ is considerably lower than the one needed to set the number of clusters used by $k$-means, for instance. Such skill makes OPF pretty much attractive to the application addressed in this paper, as discussed in the next section.



\section{Barrett's Esophagus}
\label{s.be}
\textcolor{black}{The BE disease is known as the replacement of squamous cells by columnar cells in the esophagus. This process is a result of a complication of gastroesophageal reflux disease, being able to progress into esophageal cancer~\cite{Dent2011,Sharma2016}. }

\textcolor{black}{The incidence of BE and Barrett's adenocarcinoma in the western population of the world has risen significantly in the past decade. Their close association with the metabolic syndrome suggest growth in the next years \cite{Lagergrenc6280,Dent2011,Lepage2008}. The early diagnosis of Esophageal adenocarcinoma in BE diagnosed patients is critical for remission and justifies the necessity of robust surveillance, detection, and characterization. However, the detection of dysplastic tissues and their characterization of abnormalities within BE-diagnosed patients can be challenging, especially for manual evaluation made by endoscopists. Even considering the dangerousness of the disease, when detected at the early stages, the disease can be treated with very high rates of remission (93\% after 10 years)~\cite{Dent2011,Sharma2016,Phoa555}.}

\begin{figure}[!htb]
\begin{center}
 \begin{tabular}{cc}
  \includegraphics[scale=0.11]{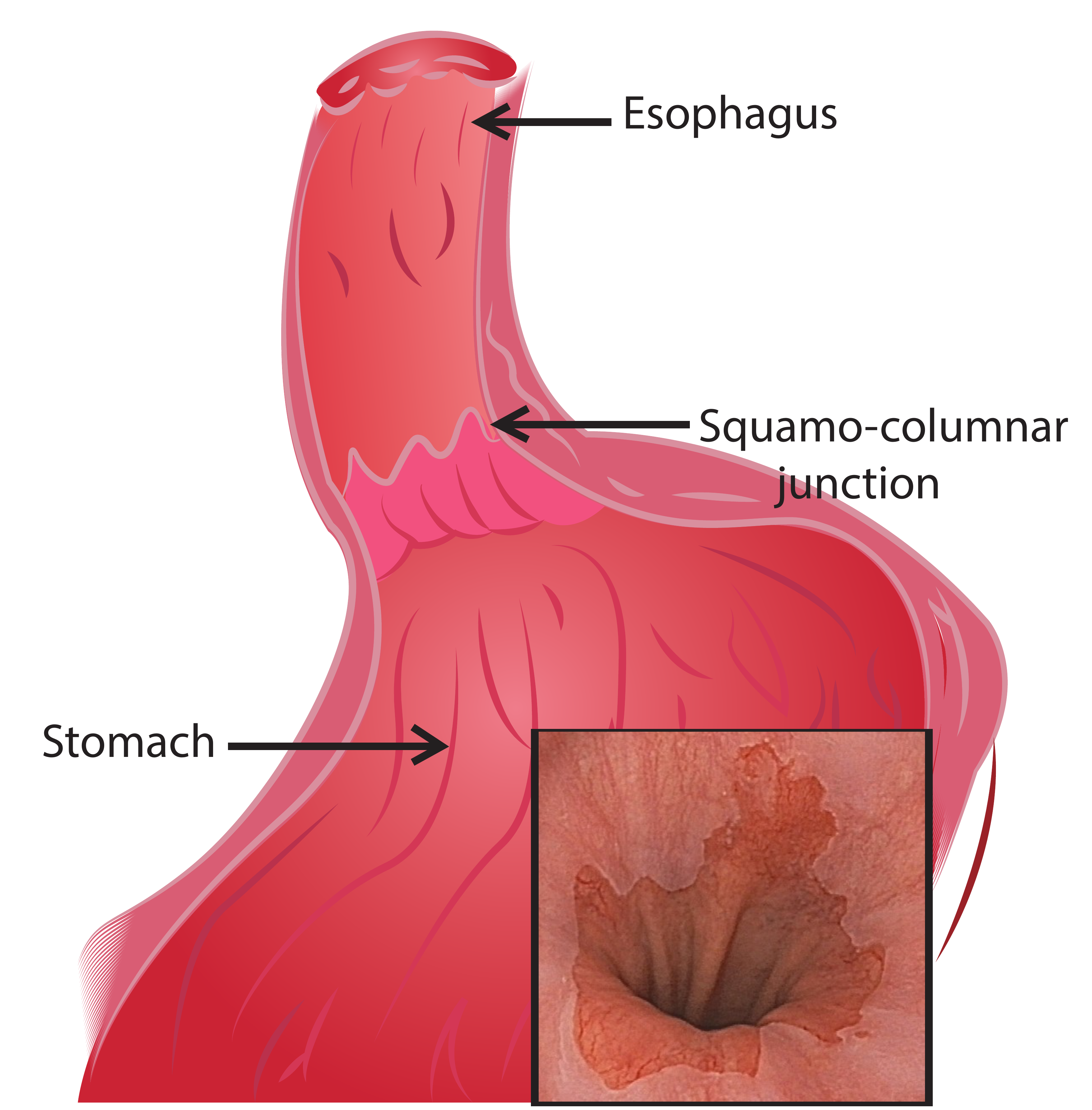} &
  \includegraphics[scale=0.11]{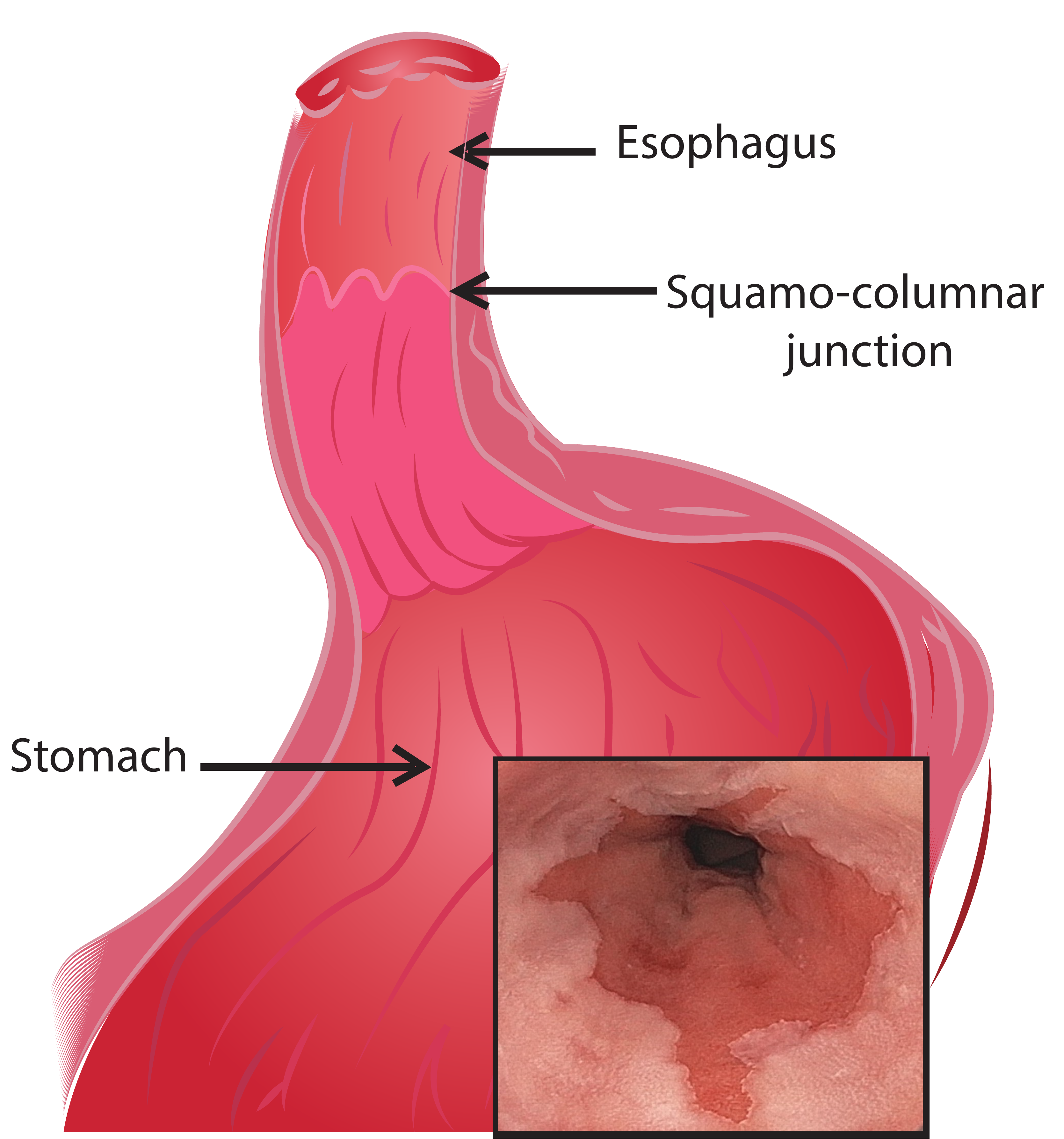}
  \\
  (a) & (b)
 \end{tabular}
\end{center}
 \caption{BE's short-segment (a) and BE's long-segment (b), with their respective endoscopic views (extracted from ~\cite{SouzaJr2018203}).} 
 \label{f.fig3}
\end{figure}

\textcolor{black}{The esophagus mucosa is composed of squamous cells (similar to the skin or mouth cells), with a whitish-pink color surface, while the gastric mucosa goes sharply from salmon-pink to red \cite{Dent2011,Sharma2016}. The point in which the stomach and the stomach meet is called squamocolumnar junction or ``Z-line". BE's mucosa may extend upward in a continuous pattern, changing the Z-line position~\cite{Dent2011,Sharma2016,Phoa555}. Figure~\ref{f.fig3} shows the two cases in which patients can present long-segment of BE and short-segment of BE in a Z-line variation.}

\section{Methodology and Proposed Approach}
\label{s.methodology}

In this section, we present the proposed approach and the methodology adopted to cope with the problem of Barrett's esophagus automatic identification using bag-of-visual-words. \textcolor{black}{First, the proposed method is defined, followed by the datasets used for the experiments, adopted classifiers and experimental delineation.}

\subsection{Proposed Method}
\label{s.method}
As mentioned earlier, one of the leading contributions of this work is to evaluate the robustness of the OPF clustering for learning visual dictionaries. To fulfill that purpose, we considered three distinct feature descriptors based on key point extraction from images: (i) SIFT, (ii) SURF, and (iii) A-KAZE. Although any other approaches could be used, we opted to employ these mainly because they are well known and widely considered in the literature of bag-of-visual-words for both image classification and retrieval, \textcolor{black}{but any other techniques could be applied considering the generalization of the learning visual dictionaries.}

\textcolor{black}{For the inicial step of the proposed model, given a set of training images, it is needed to build a bag of key points extracted from them. In hands of a feature extraction technique for the description of an image (being SURF, SIFT or A-KAZE, in this specific case), the model aims to provide the most discriminative key points in the feature dimension based on the entire feature domain. Therefore, taking into account the key points of the entire dataset, a clustering algorithm can be used to group the key points into clusters that share similar properties for choosing the ``best key point"\  from each cluster and use it as the representative of that group. Such samples will compose the final bag-of-visual-words. The main contribution of this work is the calculation of such most representative key points from clusters (as we use to call ``prototypes") by using the OPF clustering technique. After obtaining the bag-of-visual-words, the last step of the model is known as ``quantization"\, and computes the new representation for both training and testing images. For each image, it is computed the frequency of each visual word from the bag in the given image by finding the most similar visual word to each key point based on a distance metric. The outcome of that process is a histogram (feature vector) where each bin has the number of key points that are similar to its corresponding visual word. Notice that the representation of both training and testing images are computed based on the same bag. Finally, in hands of the feature vectors, each image of the evaluated dataset shows the exact same number of features for its description, but with the calculation based on the entire feature space domain. The training and testing may be conducted as the final step of the model.}  

In this work, we propose to cluster the dataset of key points using the OPF technique presented in the previous section and then use the prototypes to compose the bag-of-visual-words. As aforementioned, the prototypes are located in the regions of highest densities, which means they are pretty much suitable to describe the clusters. Another decisive point about OPF concerning other optimization-based clustering techniques relates the fact of not being attracted to local optima, such as $k$-means or $k$-medoids, for instance, which are widely used for learning dictionaries due to their simplicity and low computational cost.

\begin{figure*}[!htb]
\centering
 \begin{tabular}{c}
  \includegraphics[scale=0.4]{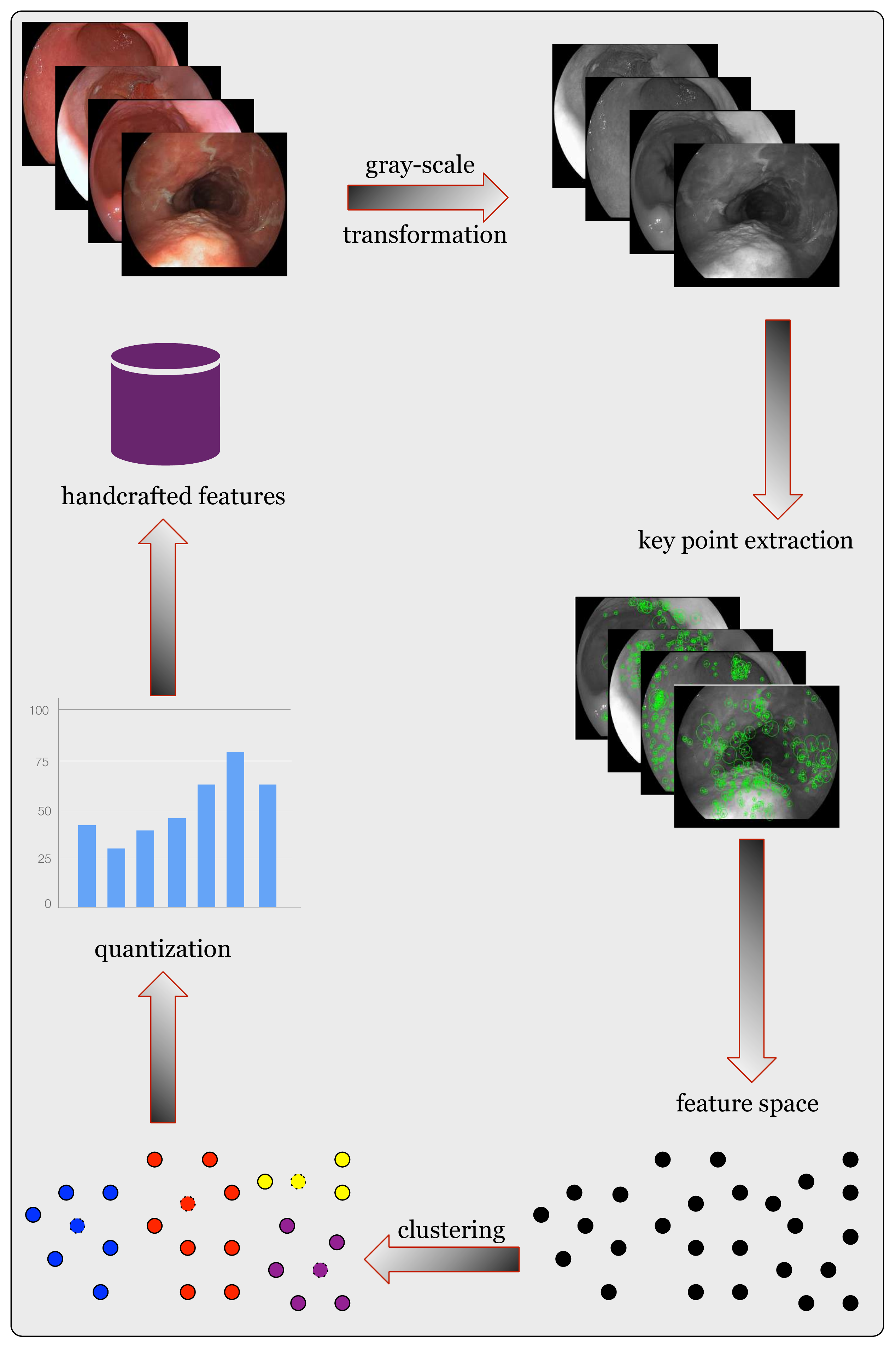} \\
 \end{tabular}
\caption{Pipeline adopted in this work for Barrett's esophagus identification.} 
 \label{s.pipeline}
\end{figure*}

As mentioned earlier, OPF finds the clusters on-the-fly, i.e., the clustering process is dynamic, and the forest configuration can change until the last sample finishes the conquering process. Instead of varying the size of the dictionary, one can change the value of $k_{max}$ and then may find the different number of clusters. \textcolor{black}{The cluster calculation comprises one of the most important steps of the proposed method. The prototype computation is performed in an unsupervised process, turning the calculation of the feature vectors based only on the key points themselves, and providing a high generalization for this task. The problem of a different number of key points for each image can be solved using this bag-of-visual-words approach, proposing a consistent way of regular description for images evaluated by feature extraction techniques.} Figure~\ref{s.pipeline} depicts the pipeline adopted in this work.

Since the images are colored, a gray-scale normalization is applied to the images so that the key points can be extracted. Later, such PoIs are then mapped onto a feature space for clustering purposes. An example of the outcome of the clustering process is depicted at the bottom of Figure~\ref{s.pipeline}. Each color stands for a different group and the dashed nodes represent the prototypes selected by OPF to be part of the visual dictionary. \textcolor{black}{As aforementioned, a histogram is built upon the training PoIs and the visual words for the further design of the final set of handcrafted features. For the selected visual words, an evaluation of their appearance is performed in the PoIs of each dataset image aiming to calculate the final cumulative histogram that represents each feature vector, with dimension depending on the number of visual words generated in the clustering calculation of the bag.}

\subsection{Datasets}
\label{s.datasets}

An in-depth analysis concerning the robustness of the proposed approach is provided through two datasets. The first dataset comprises a set of images from a benchmark dataset provided at the ``MICCAI 2015 EndoVis Challenge"\footnote{\url{https://endovissub-barrett.grand-challenge.org/home/}} was considered, hereinafter called ``MICCAI 2015"\ dataset, which aimed at differentiating Barrett's esophagus from cancerous images. Such dataset is composed of $100$ endoscopic pictures of the lower esophagus captured from $39$ individuals, $22$ of them being diagnosed with early-stage Barrett's, and $17$ showing signs of esophageal adenocarcinoma. Each patient has several endoscopic images available, ranging from one to a maximum of eight. The database comprises a total of $50$ images displaying cancerous tissue areas as well as $50$ images showing dysplasia without signs of cancer. Suspicious lesions observed in the cancerous images had been delineated individually by five endoscopy experts.

Additionally, a dataset provided by the Augsburg Klinikum, Medizinische Klinik III was also used for the experiments. Such dataset is composed of $76$  endoscopic images (esophagus) captured from different patients with adenocarcinoma ($34$ samples) and BE ($42$ samples). The images were annotated (manual segmentation of the adenocarcinoma's and Barrett's area, respectively) by an expert from the Augsburg Klinikum, and the diagnosis was provided using biopsy. Since we are dealing with a classification problem, the annotations provided by the experts were not considered in our work.

\begin{figure*}[!htb]
\centering
 \begin{tabular}{c}
  \includegraphics[scale=0.55]{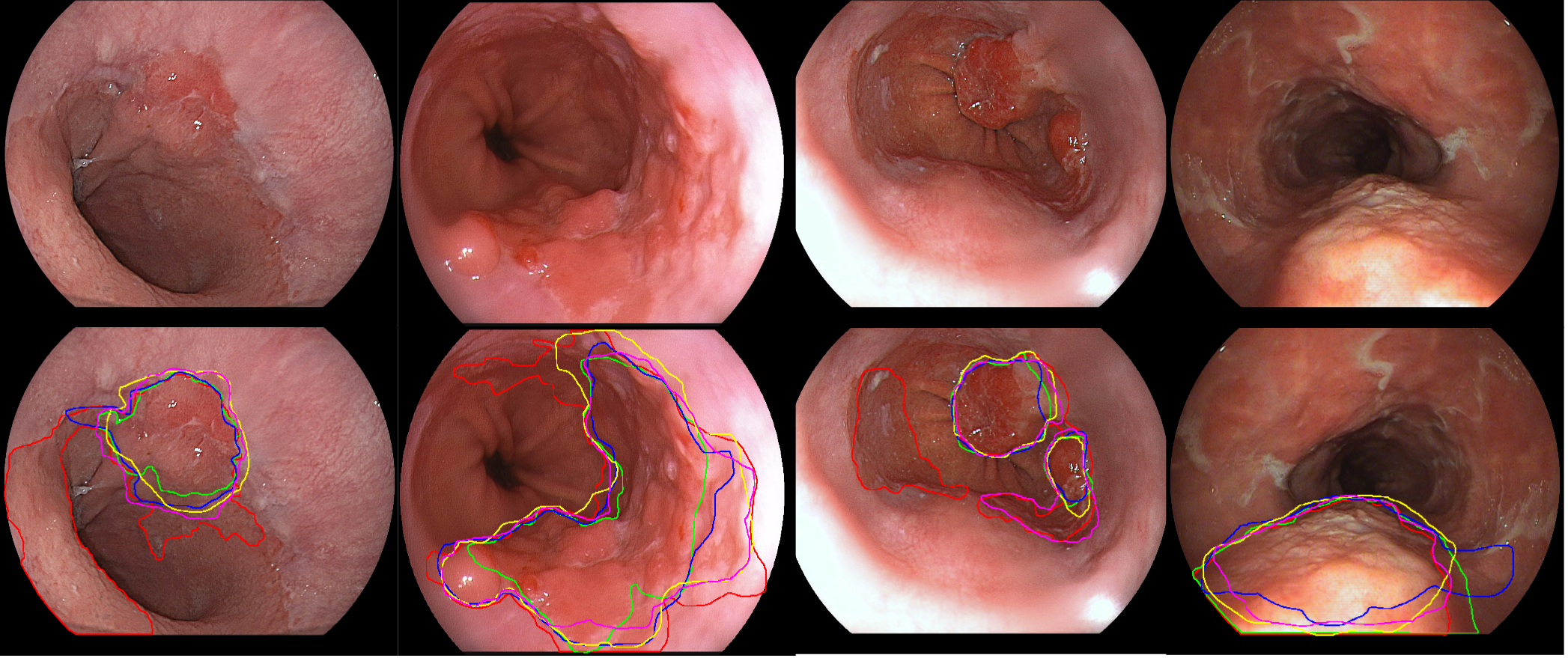} \\
 \end{tabular}
\caption{Some examples of images positive for cancer and their respective delineations (MICCAI 2015 dataset).} 
 \label{f.MICCAI}
\end{figure*}

Figure~\ref{f.MICCAI} depicts some examples of the MICCAI 2015 dataset positive for cancer (i.e., negative for BE) and their respective delineations performed by five experts. However, we are not working with the delineation information since we compute the PoIs for the whole image. One could use the information about the delineated regions to extract PoIs from that areas only, which could guarantee that pure adenocarcinoma PoIs are computed, but the problem still concerns the fact that delineations are not available to all real-world images.

\begin{figure*}[!htb]
\centering
 \begin{tabular}{c}
  \includegraphics[scale=0.54]{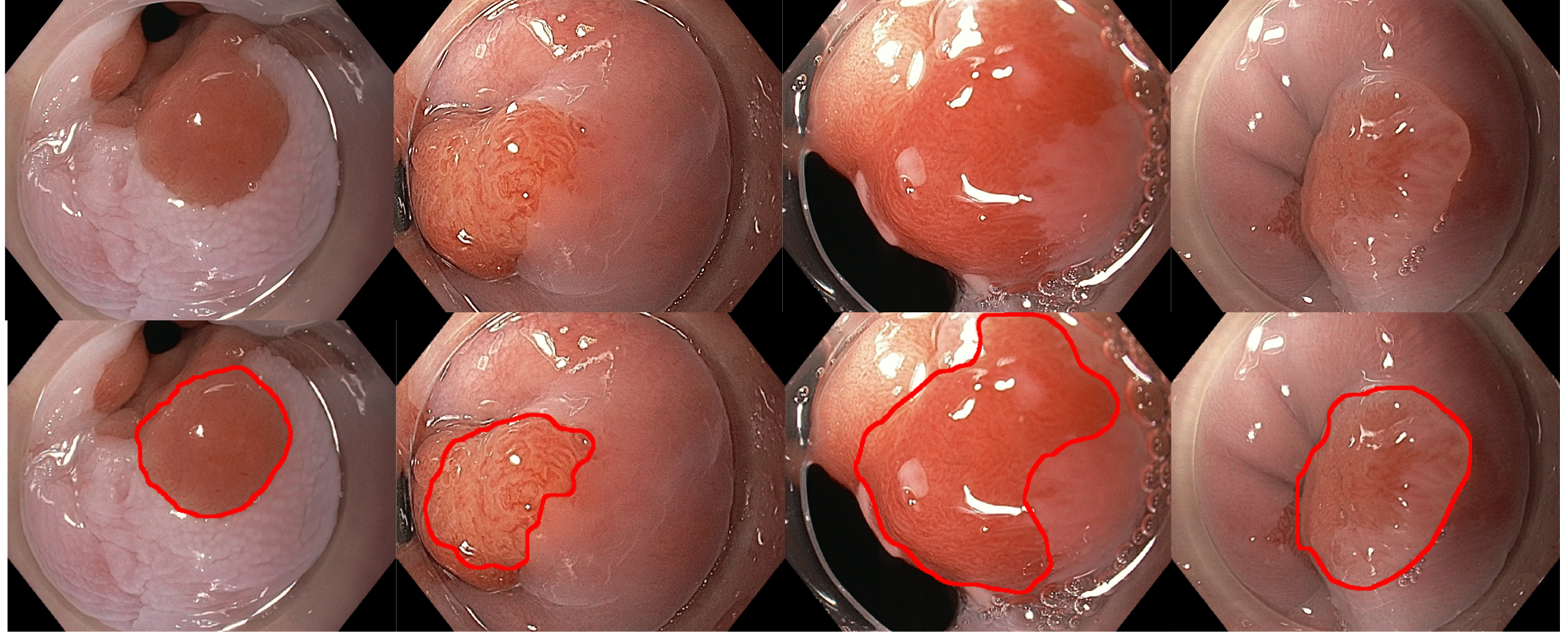} \\
 \end{tabular}
 \caption{Some examples of images positive for cancer and their respective delineations (Augsburg dataset).} 
 \label{f.Augsburg}
\end{figure*}

Figure~\ref{f.Augsburg} displays some images positive for cancer from Augsburg dataset. In this case, we have only one delineation per image. Once again, such information is not used in this work since we are interested mostly in the differentiation of Barrett's esophagus and adenocarcinoma rather than its segmentation.

\subsection{Adopted Classifiers}
\label{s.classifiers}
We considered different supervised pattern recognition techniques to assess the robustness of unsupervised OPF for learning visual dictionaries:

\begin{itemize}
\item OPF$_{cpl}$: supervised OPF with complete graph proposed by Papa et al.~\cite{PapaIJIST:09,PapaPR:12};
\item OPF$_{knn}$: supervised OPF with $k$-nn graph proposed by Papa et al.~\cite{PapaPRL:17};
\item SVM-RBF: Support Vector Machines with Radial Basis Function kernel and parameters optimized by cross-validation~\cite{ChangACMTIST:11};
\item \begin{sloppypar}SVM-Linear: Support Vector Machines with Linear kernel and parameters optimized by cross-validation~\cite{ChangACMTIST:11};\end{sloppypar} 
\item Bayes: standard Bayesian classifier.
\end{itemize}
Regarding the OPF-based classifiers, we used the LibOPF~\cite{LibOPF:14}, which is an open-source library that implements both the supervised as well as the unsupervised versions of the OPF used in this work. With respect to the Bayesian classifier, we employed our own implementation.

\subsection{Experimental Delineation}
\label{s.delineation}

To compose the set of experiments, we considered three different sizes for the dictionaries: $100$, $500$, and $1,000$. The main idea is to evaluate the robustness of the techniques used in this work under different scenarios. As we shall discuss later, the usage of dictionaries with $500$ visual words seemed to achieve better results, as stated in a previous work~\cite{Souza_SIBGRAPI:17}, which motivated us to set $k_{max}=500$ for this one. However, this not implies in constraining OPF to find exactly $500$ clusters, just to limit the size of the neighborhood of each sample to be $500$. Regarding OPF$_{knn}$, its parameter $k$ is fine-tuned within the range $[1,500]$, and the value that maximized the accuracy over the training set was used.

Regarding the experimental validation, it was considered a cross-validation approach with $20$ runs and using $70\%$ of the dataset for training purposes, as well as the remaining $30\%$ for classification. Moreover, the experimental results were assessed using a statistical analysis using the Wilcoxon signed-rank test with confidence as of $5\%$~\cite{Wilcoxon:45}. All experiments were conducted on an $8$GB-memory computer equipped with an Intel Core i5 - 2.30 GHz processor. Additionally, we employed the OpenCV~\cite{OpenCV:15} implementation for feature extraction using SIFT, SURF, and A-KAZE.



\section{Experimental Results}
\label{s.experiments}

In this section, we present the experiments used to evaluate the proposed approach. Five supervised classifiers were considered to discriminate between samples positive and negative to adenocarcinoma: OPF$_{cpl}$, OPF$_{knn}$, SVM-RBF, SVM-Linear, and Bayesian classifier (hereinafter called Bayes). \textcolor{black}{For all the classifiers adopted for such evaluation, there was no need for setting any parameter, as long as they were used in the default set.} The same experimental protocol was applied to all techniques using cross-validation, i.e., three distinct feature representations were considered (SURF, SIFT, and A-KAZE, \textcolor{black}{with the metric threshold of all sets is default}), and with different bag sizes (i.e., $100$, $500$ and $1,000$ visual words). The results are presented and discussed considering each dataset individually.

\begin{sloppypar}A statistical evaluation using the signed-rank Wilcoxon test~\cite{Wilcoxon:45} was used for comparison purposes as follows:\end{sloppypar}

\begin{enumerate}
\item For each dictionary generation approach (i.e., clustering by $k$-means, random or unsupervised OPF), it was verified the classification results and the best ones were highlighted in bold. Statistically similar results were highlighted in bold.
\item For each feature extractor (i.e., A-KAZE, SIFT, and SURF), the best statistical results were underlined.
\item Additionally, the best results among all configurations were marked with a `$\star$'\ symbol.
\end{enumerate}
This very same procedure was adopted to both datasets. 

In this work, we used the following accuracy rate:
\begin{equation}
\label{e.recognition_rate}
A=\frac{TP+TN}{TP+TN+FP+FN}\cdot100,
\end{equation}
where $TP$ and $TN$ stand for the true positives and true negatives, respectively, and $FN$ and $FP$ denote the false negatives and false positives, respectively. In a nutshell, the above equation computes the ratio between the number of correct classifications (i.e., $TP+TN$) and the size of the dataset (i.e., all correct and wrong classifications).

\subsection{MICCAI 2015 Dataset}
\label{ss.miccai}

Tables~\ref{t.akaze1},~\ref{t.surf1}, and~\ref{t.sift1} present the results related to A-KAZE, SURF, and SIFT descriptors, respectively, concerning MICCAI 2015 dataset. Regarding the A-KAZE results presented in Table~\ref{t.akaze1}, one can draw the following conclusions: (i) OPF$_{cpl}$ obtained the best results for all dictionary generation techniques, and (ii) OPF clustering achieved the best results ($77.6$\% of recognition rate with $1,000$ visual words) for BE recognition among all configurations, although being statistically similar to $k$-means with OPF$_{cpl}$ with $500$ and $1,000$ visual words as well.

\begin{table}[!htb]
\caption{Mean accuracy results using A-KAZE features with $100$, $500$, and $1,000$ visual words.}
\begin{center}
\resizebox{0.48\textwidth}{!}{%
\begin{tabular}{cc|c|c|c}
    \hline
    \multicolumn{2}{c|}{Dictionary} &100& 500 & 1000\\
    \hline
    \multirow{3}{*}{$k$-means} & OPF$_{cpl}$& 73.6\% & \textbf{\underline{$^{\star}$76.3\%}} & \textbf{\underline{$^{\star}$77.2\%}}\\
	\multirow{3}{*}{}& OPF$_{knn}$ & 59.2\% & 61.7\% & 68.1\%\\
    \multirow{3}{*}{}& SVM-RBF & 60.7\% & 65.9\% & 66.1\%\\
    \multirow{3}{*}{} & SVM-Linear& 58.5\% & 63.0\% & 67.4\%\\
    \multirow{3}{*}{} & Bayes& 56.8\% & 60.0\% & 60.9\%\\
    \hline
     \multirow{3}{*}{Random} & OPF$_{cpl}$& 59.5\% & 63.9\% & \textbf{70.3\%}\\
	 \multirow{3}{*}{}& OPF$_{knn}$ & 58.3\% & 61.7\% & 62.3\%\\
    \multirow{3}{*}{}& SVM-RBF & 62.1\% & 65.6\% & 63.7\%\\
    \multirow{3}{*}{} & SVM-Linear & 55.3\% & 59.0\% & 59.1\%\\
    \multirow{3}{*}{} & Bayes& 55.5\% & 62.2\% & 61.1\%\\
    \hline
     \multirow{3}{*}{OPF clustering} & OPF$_{cpl}$& 72.2\% & 73.1\% & \textbf{\underline{$^{\star}$77.6\%}}\\
	 \multirow{3}{*}{}& OPF$_{knn}$ & 62.3\% & 60.1\% & 66.2\%\\
    \multirow{3}{*}{}& SVM-RBF & 61.9\% & 65.1\% & 70.9\%\\
    \multirow{3}{*}{} & SVM-Linear& 55.8\% & 60.5\% & 66.8\%\\
    \multirow{3}{*}{} & Bayes& 55.8\% & 58.0\% & 61.3\%\\
    \hline
\end{tabular}
}
\end{center}
\label{t.akaze1}
\end{table}

The average number of PoIs used for training and test sets concerning A-KAZE feature extractor were $16,024$ and $6,868$, respectively, taking an average computational load of $4.05$ minutes. A training set composed of around $16,000$ visual words is enough to support the sizes of the dictionaries we used to build the feature vector of each image, i.e., $100$, $500$, $1,000$. Larger dictionaries may not be interesting since there will be numerous small-sized clusters, which means less spatial information about the visual words is captured.

Table~\ref{t.surf1} presents the results concerning the SURF feature extractor. Once again, OPF$_{cpl}$ achieved the best classification results regarding all dictionary generation approaches, and OPF clustering allowed the best results among all, i.e., it could learn better dictionaries for image representation. In this context, a dictionary of size $500$ computed by $k$-means also achieved the best recognition rates according to the statistical test. The average number of PoIs used for training and test sets concerning SURF feature extractor were $14,411$ and $6,189$, respectively, taking an average computational load of $13.77$ minutes.

\begin{table}[!htb]
\caption{Mean accuracy results using SURF Features and $100$, $500$, and $1,000$ visual words.}
\begin{center}
\resizebox{0.48\textwidth}{!}{%
\begin{tabular}{cc|c|c|c}
    \hline
    \multicolumn{2}{c|}{Dictionary} & 100 & 500 & 1000\\
    \hline
    \multirow{3}{*}{$k$-means} & OPF$_{cpl}$& 70.0\% & \textbf{\underline{74.8\%}} & \textbf{73.6\%}\\
	\multirow{3}{*}{}& OPF$_{knn}$ & 64.1\% & 66.0\% & 65.1\%\\
    \multirow{3}{*}{}& SVM-RBF & 63.6\% & 64.8\% & 62.6\%\\
    \multirow{3}{*}{} & SVM-Linear & 62.0\% & 58.6\% & 62.8\%\\
    \multirow{3}{*}{} & Bayes & 56.4\% & 56.9\% & 57.4\%\\
    \hline
     \multirow{3}{*}{Random} & OPF$_{cpl}$& \textbf{69.7\%} & \textbf{70.2\%}  & \textbf{66.1\%}\\
	 \multirow{3}{*}{}& OPF$_{knn}$ & 58.0\% & 58.4\% & 61.8\%\\
    \multirow{3}{*}{}& SVM-RBF & 61.0\% & 63.4\% & 62.1\%\\
    \multirow{3}{*}{} & SVM-Linear & 51.7\% & 57.6\% & 56.5\%\\
    \multirow{3}{*}{} & Bayes& 50.5\% & 53.5\% & 56.9\%\\
    \hline
     \multirow{3}{*}{OPF clustering} & OPF$_{cpl}$& 69.4\% & \textbf{\underline{$^{\star}$78.4\%}}  & \textbf{\underline{$^{\star}$77.1\%}}\\
	 \multirow{3}{*}{}& OPF$_{knn}$ & 63.6\% & 69.6\% & 71.6\%\\
    \multirow{3}{*}{}& SVM-RBF & 67.5\% & 71.8\% & 70.9\%\\
    \multirow{3}{*}{} & SVM-Linear & 65.1\% & 66.9\% & 66.7\%\\
    \multirow{3}{*}{} & Bayes& 53.3\% & 56.8\% & 57.1\%\\
    \hline
\end{tabular}
}
\end{center}
\label{t.surf1}
\end{table}

One can observe that SVM did not obtain proper recognition rates in both situations, i.e., A-KAZE and SURF feature extractors. One possible reason concerns the number of training samples, which is usually lower than the number of features. Therefore, SVM will map samples to a lower-dimensionality feature space instead of a higher one, thus neglecting the assumption of linearity in higher-dimensionality spaces.

Table~\ref{t.sift1} presents the results considering the SIFT feature extractor. Once again, OPF$_{cpl}$ achieved the best results so far, with OPF$_{knn}$ and SVM-RBF being statistically similar for $k$-means with $1,000$ words and a random generation of dictionaries with $1,000$ words. However, the best global results were achieved using OPF clustering with OPF$_{cpl}$ with $500$ and $1,000$ visual words, outperforming by far the other results with SIFT feature extractor. The average number of PoIs used for training and test sets concerning SIFT feature extractor were $28,137$ and $12,059$, respectively, taking an average computational load of $5.95$ minutes.

\begin{table}[h!]
\caption{Mean accuracy results using SIFT Features and $100$, $500$, and $1,000$ visual words.}
\begin{center}
\resizebox{0.48\textwidth}{!}{%
\begin{tabular}{cc|c|c|c}
    \hline
    \multicolumn{2}{c|}{Dictionary} &100& 500 & 1000\\
    \hline
    \multirow{3}{*}{$k$-means} & OPF$_{cpl}$& 68.3\% & \textbf{72.3\%} & \textbf{71.4\%}\\
	\multirow{3}{*}{}& OPF$_{knn}$ & 67.0\% & \textbf{71.8\%} & \textbf{72.1\%}\\
    \multirow{3}{*}{}& SVM-RBF & 67.3\% & \textbf{71.4\%} & \textbf{71.9\%}\\
    \multirow{3}{*}{} & SVM-Linear & 55.2\% & 56.8\% & 67.3\%\\
    \multirow{3}{*}{} & Bayes& 53.5\% & 60.0\% & 60.7\%\\
    \hline
     \multirow{3}{*}{Random} & OPF$_{cpl}$& \textbf{66.4\%} & \textbf{70.7\%} & \textbf{71.2\%}\\
	 \multirow{3}{*}{}& OPF$_{knn}$ & 58.1\% & 63.9\% & \textbf{66.1\%}\\
    \multirow{3}{*}{}& SVM-RBF & 62.1\% & 65.6\% & 63.7\%\\
    \multirow{3}{*}{} & SVM-Linear & 53.2\% & 54.5\% & 52.7\%\\
    \multirow{3}{*}{} & Bayes& 50.2\% & 53.0\% & 54.4\%\\
    \hline
     \multirow{3}{*}{OPF clustering} & OPF$_{cpl}$& 71.2\% & \textbf{\underline{$^{\star}$77.7\%}} & \textbf{\underline{$^{\star}$78.9\%}}\\
	 \multirow{3}{*}{}& OPF$_{knn}$ & 63.9\% & 71.3\% & \textbf{75.7\%}\\
    \multirow{3}{*}{}& SVM-RBF & 68.0\% & 70.2\% & 69.7\%\\
    \multirow{3}{*}{} & SVM-Linear & 61.3\% & 64.7\% & 64.4\%\\
    \multirow{3}{*}{} & Bayes& 50.2\% & 53.0\% & 54.4\%\\
    \hline
\end{tabular}
}
\end{center}
\label{t.sift1}
\end{table}

Last but not least, the best results among all three feature extractors (i.e., the ones marked with `$\star$') were obtained using OPF clustering for dictionary generation and OPF$_{cpl}$ for classification with $500$ and $1,000$ visual words considering SURF and SIFT, and the same pair (i.e., OPF clustering and OPF$_{cpl}$) regarding A-KAZE with $1,000$ visual words, and finally $k$-means and OPF$_{cpl}$ with $500$ words. Notice the best absolute result was obtained using OPF clustering for visual words generation and OPF$_{cpl}$ for classification purposes with SIFT-based features (i.e., $78.9\%$).

\subsection{Augsburg Dataset}
\label{ss.augsburg}

Tables~\ref{t.akaze2},~\ref{t.surf2}, and~\ref{t.sift2} present the results related to A-KAZE, SURF, and SIFT descriptors, respectively, concerning Augsburg dataset. Starting with the A-KAZE feature extractor, one can observe the best results were mostly obtained by both OPF$_{cpl}$ and OPF$_{knn}$. The best global results were achieved by OPF clustering, Random and $k$-means, but the most accurate one (i.e., absolute results) was OPF clustering for visual dictionary generation and OPF$_{cpl}$ for classification purposes with accuracy of $72.6\%$. Such result is slightly less accurate than the same feature extractor considering MICCAI 2015 dataset since Augsburg dataset is more challenging due to different levels of adenocarcinoma. The average number of PoIs used for training and test sets concerning A-KAZE feature extractor were $40,064$ and $17,170$, respectively, taking an average computational load of $4.18$ minutes.

\begin{table}[!htb]
\caption{Mean accuracy results using A-KAZE Features and $100$, $500$, and $1,000$ visual words.}
\begin{center}
\resizebox{0.48\textwidth}{!}{%
\begin{tabular}{cc|c|c|c}
    \hline
    \multicolumn{2}{c|}{Dictionary} &100& 500 & 1000\\
    \hline
    \multirow{3}{*}{$k$-means} & OPF$_{cpl}$& 60.7\% & \textbf{\underline{$^{\star}$69.4\%}} & 65.6\%\\
	\multirow{3}{*}{}& OPF$_{knn}$ & 61.9\% & \textbf{66.1\%} & \textbf{\underline{$^{\star}$70.1\%}}\\
    \multirow{3}{*}{}& SVM-RBF & 60.4\% & 63.5\% & 63.1\%\\
    \multirow{3}{*}{} & SVM-Linear & 55.1\% & 60.4\% & 62.1\%\\
    \multirow{3}{*}{} & Bayes& 56.9\% & 60.1\% & 61.3\%\\
    \hline
     \multirow{3}{*}{Random} & OPF$_{cpl}$& 59.4\% & \textbf{68.4\%} & \textbf{\underline{$^{\star}$69.9\%}}\\
	 \multirow{3}{*}{}& OPF$_{knn}$ & 59.9\% & 61.4\% & 62.4\%\\
    \multirow{3}{*}{}& SVM-RBF & 57.9\% & 62.2\% & 63.2\%\\
    \multirow{3}{*}{} & SVM-Linear & 55.3\% & 58.8\% & 58.9\%\\
    \multirow{3}{*}{} & Bayes& 56.5\% & 57.1\% & 61.0\%\\
    \hline
     \multirow{3}{*}{OPF clustering} & OPF$_{cpl}$& 68.4\% & \textbf{\underline{$^{\star}$68.7\%}} & \textbf{\underline{$^{\star}$72.6\%}}\\
	 \multirow{3}{*}{}& OPF$_{knn}$ & 67.4\% & \textbf{\underline{$^{\star}$69.3\%}} & \textbf{\underline{$^{\star}$70.3\%}}\\
    \multirow{3}{*}{}& SVM-RBF & 59.4\% & 63.0\% & \textbf{\underline{$^{\star}$69.8\%}}\\
    \multirow{3}{*}{} & SVM-Linear & 57.7\% & 57.3\% & 62.7\%\\
    \multirow{3}{*}{} & Bayes& 62.4\% & 60.7\% & 63.1\%\\
    \hline
\end{tabular}
}
\end{center}
\label{t.akaze2}
\end{table}

Table~\ref{t.surf2} presents the results concerning the SURF feature extractor. Once again, OPF-based classifiers obtained the best results in most of the scenarios, being OPF clustering and $k$-means the best approaches for visual dictionary generation. The best absolute classification results were obtained by OPF$_{cpl}$ and Bayes with accuracies nearly to $68\%$. The average number of PoIs used for training and test sets concerning SURF feature extractor were $14,251$ and $6,108$, respectively, taking an average computational load of $9.23$ minutes.

\begin{table}[!htb]
\caption{Mean accuracy results using SURF Features and $100$, $500$, and $1,000$ words.}
\begin{center}
\resizebox{0.48\textwidth}{!}{%
\begin{tabular}{cc|c|c|c}
    \hline
    \multicolumn{2}{c|}{Dictionary} & 100 & 500 & 1000\\
    \hline
    \multirow{3}{*}{$k$-means} & OPF$_{cpl}$& \textbf{\underline{66.3\%}} & \textbf{\underline{$^{\star}$67.9\%}} & 61.5\%\\
	\multirow{3}{*}{}& OPF$_{knn}$ & 62.8\% & 63.2\% & \textbf{\underline{65.4\%}}\\
    \multirow{3}{*}{}& SVM-RBF & 57.1\% & 61.1\% & 62.9\%\\
    \multirow{3}{*}{} & SVM-Linear & 56.7\% & 57.1\% & 59.4\%\\
    \multirow{3}{*}{} & Bayes& 60.8\% & 59.9\% & 61.1\%\\
    \hline
     \multirow{3}{*}{Random} & OPF$_{cpl}$& 60.0\% & \textbf{62.2\%}  & \textbf{63.5\%}\\
	 \multirow{3}{*}{}& OPF$_{knn}$ & 54.2\% & 58.1\% & 60.8\%\\
    \multirow{3}{*}{}& SVM-RBF & 61.3\% & \textbf{61.9\%} & \textbf{62.0\%}\\
    \multirow{3}{*}{} & SVM-Linear & 57.1\% & 55.4\% & 56.4\%\\
    \multirow{3}{*}{} & Bayes& 51.9\% & 59.0\% & 59.1\%\\
    \hline
     \multirow{3}{*}{OPF clustering} & OPF$_{cpl}$& 59.2\% & 62.1\%  & \textbf{\underline{66.1\%}}\\
	 \multirow{3}{*}{}& OPF$_{knn}$ & 61.1\% & 63.9\% & 64.5\%\\
    \multirow{3}{*}{}& SVM-RBF & 58.5\% & 62.0\% &\textbf{\underline{65.4\%}}\\
    \multirow{3}{*}{} & SVM-Linear & 53.5\% & 60.8\% & 64.6\%\\
    \multirow{3}{*}{} & Bayes& 59.8\% & \textbf{\underline{67.0\%}} & \textbf{\underline{$^{\star}$67.9\%}}\\
    \hline
\end{tabular}
}
\end{center}
\label{t.surf2}
\end{table}

The Augsburg dataset figured out as being more challenging than MICCAI 2015 dataset due to the considerably low results achieved (Table~\ref{t.surf1}). SVM-RBF presented better results with higher-dimensionality bags (i.e., $65.4\%$ with $1,000$ words with OPF clustering), and the same behavior can be observed regarding SVM-Linear.

Table~\ref{t.sift2} presents the results considering the SIFT feature extractor. In this case, OPF-based classifiers and SVM-RBF figured as the most accurate techniques and OPF clustering as the best one for visual dictionary generation (absolute results). A comparison against A-KAZE and SURF showed these to be quite more accurate than SIFT, an opposite situation that occurred over MICCAI 2015 dataset, where SIFT achieved the best recognition rates. Additionally, the average number of PoIs used for training and test sets concerning SIFT feature extractor were $89,514$ and $38,363$, respectively, taking an average computational load of $8.71$ minutes.

\begin{table}[!htb]
\caption{Mean accuracy results using SIFT Features and $100$, $500$, and $1,000$ visual words.}
\begin{center}
\resizebox{0.48\textwidth}{!}{%
\begin{tabular}{cc|c|c|c}
    \hline
    \multicolumn{2}{c|}{Dictionary} &100& 500 & 1000\\
    \hline
    \multirow{3}{*}{$k$-means} & OPF$_{cpl}$& \textbf{60.3\%} & \textbf{60.5\%} & 59.3\%\\
	\multirow{3}{*}{}& OPF$_{knn}$ & 58.9\% & \textbf{60.6\%} & \textbf{\underline{62.0\%}}\\
    \multirow{3}{*}{}& SVM-RBF & 60.8\% & 61.8\% & \textbf{59.8\%}\\
    \multirow{3}{*}{} & SVM-Linear & 55.5\% & 57.1\% & \textbf{59.9\%}\\
    \multirow{3}{*}{} & Bayes& 53.1\% & 54.8\% & 58.7\%\\
    \hline
     \multirow{3}{*}{Random} & OPF$_{cpl}$& 59.2\% & 60.5\% & \textbf{61.6\%}\\
	 \multirow{3}{*}{}& OPF$_{knn}$ & 57.0\% & 58.4\% & 60.5\%\\
    \multirow{3}{*}{}& SVM-RBF & 57.8\% & \textbf{\underline{62.6\%}} & \textbf{\underline{62.1\%}}\\
    \multirow{3}{*}{} & SVM-Linear & 54.4\% & 55.6\% & \textbf{61.5\%}\\
    \multirow{3}{*}{} & Bayes& 51.9\% & 57.0\% & 59.0\%\\
    \hline
     \multirow{3}{*}{OPF clustering} & OPF$_{cpl}$& 60.4\% & \textbf{\underline{62.8\%}} & \textbf{\underline{62.1\%}}\\
	 \multirow{3}{*}{}& OPF$_{knn}$ & 58.1\% & 61.6\% & \textbf{\underline{63.9\%}}\\
    \multirow{3}{*}{}& SVM-RBF & 57.0\% & 60.5\% & \textbf{\underline{62.1\%}}\\
    \multirow{3}{*}{} & SVM-Linear & 58.8\% & 58.9\% & 58.7\%\\
    \multirow{3}{*}{} & Bayes& 61.1\% & \textbf{\underline{62.2\%}} & 61.8\%\\
    \hline
\end{tabular}
}
\end{center}
\label{t.sift2}
\end{table}
 
\subsection{Discussion}
\label{ss.discussion}

\begin{sloppypar}In this section, we aim at providing a more in-depth discussion about the experiments, as well as insightful conclusions regarding the usage of bag-of-visual words in the context of computer-aided differentiation between Barrett's esophagus and adenocarcinoma. Table~\ref{t.summary} presents a summary with the best results obtained in the previous two sections concerning the number of visual words and feature extractor. Concerning both datasets, OPF$_{cpl}$ figured as the more accurate classification technique, meanwhile OPF clustering appears as the best dictionary generation approach.\end{sloppypar}

\begin{table}[h!]
\caption{Summarization of the results.}
\begin{center}
\resizebox{0.48\textwidth}{!}{%
\begin{tabular}{c|c|c|c}
    \hline
    \textbf{Dataset} & \textbf{Accuracy} & \textbf{Feature Extractor} & \textbf{\#visual words}\\
    \hline
    MICCAI 2015& 78.9\% & SIFT & $1,000$ \\\hline
	Augsburg & 72.6\% & A-KAZE & $1,000$\\\hline
\end{tabular}
}
\end{center}
\label{t.summary}
\end{table}

The results support the primary contributions stated previously, which are related to the robustness of OPF-based classifiers for both supervised and unsupervised learning in the context of automatic adenocarcinoma identification. Additionally, the number of visual words strongly affects the results, but we believe a trade-off between the size of the dictionary and the information it carries on shall be established beforehand.

Table~\ref{t.global} summarizes the mean sensitivity and specificity results of both datasets with the best configuration of the number of visual words, dictionary generation approach, feature extractor, and classification technique mentioned above. Sensitivity stands for the classification rate considering adenocarcinoma identification, i.e., those positive to Barrett's esophagus and to adenocarcinoma, and specificity denotes the accuracy regarding those negative to adenocarcinoma, i.e., positive only to BE. \textcolor{black}{Considering such sensitivity and specificity results, some conclusions can be drawn: (i) for the MICCAI 2015 dataset, the sensitivity results presented higher values than the specificity ones, suggesting a very good generalization in the positive adenocarcinoma identification. Even with lower results, the specificity still showed a convincing value, and the misclassification can be justified by two factors: the fuzzy region (region in which the experts disagree in the annotation) and lack of enough key points in the non-cancerous regions during the feature vector calculation. For the Augsburg results of sensitivity and specificity, a better trade-off between the correct classification of positive and non-positive adenocarcinoma samples could be found, but still with lower results when compared to the MICCAI 2015 dataset ones. The Augsburg dataset presents images with different behavior and acquisition technology when compared to the MICCAI 2015 ones, thus justifying the worse results.}

\begin{table}[h!]
\caption{Mean sensitivity (i.e., positive to BE) and specificity (i.e., negative to BE) results.}
\begin{center}
\begin{tabular}{c|c|c}
    \hline
    \textbf{Dataset} & \textbf{Sensitivity} & \textbf{Specificity}\\
    \hline
    MICCAI 2015 & 81.7\% & 76.4\%\\\hline
	Augsburg & 70.9\% & 74.9\%\\\hline
\end{tabular}
\end{center}
\label{t.global}
\end{table}

To provide more insightful comments and to better understand the working mechanism of visual words in the context of computer-assisted BE identification, we performed some additional experiments with cancerous images that were classified either as cancer or as Barrett's esophagus since we have their delineated regions. In a nutshell, the main idea is to compute the percentage of PoIs located inside those regions with respect to the remaining ones (i..e, those located outside cancerous areas). This information allows us to compare whether the number of PoIs placed inside the delineated regions are enough or not to provide accurate classifications. 

\textcolor{black}{Table~\ref{t.cancer.pois.database} presents the mean percentage of PoIs located inside the cancerous area for the whole dataset, as well as the average percentage of PoI inside the cancerous area concerning the misclassified images (i.e., cancerous images that were classified as BE). Since we conducted a cross-validation approach with $20$ runs, the average percentages concerning the misclassified images (i.e., Cancer$\rightarrow$BE) were computed to each run, for the further computation of the average value of all. Additionally, since MICCAI 2015 dataset comprises delineations from five experts, we took the intersection of them all as the final delineated area to compute the percentage of PoIs into account.}

One can observe that the percentage of PoIs inside the cancerous images were more significant than the values obtained from the misclassified images. Such assumption is pretty interesting since we can conclude that the number of PoIs inside the delineated regions are essential to achieve accurate results and to avoid misclassifications. The only exception stands for the Augsburg dataset with A-KAZE features, where the number of PoIs were slightly higher for the misclassified images. Note that the percentage of PoIs inside the cancer region is in general higher for the Augsburg databaset than for the MICCAI 2015 dataset. This can be explained because the Augsburg images use the near-focal imaging technique, in which the suspicious region is displayed larger.

\begin{table}[!htb]
\caption{Percentage of PoIs inside the delineated (cancerous) ares.}
\begin{center}
\begin{tabular}{c|c|c|c}
    \hline
    \textbf{Dataset} & \textbf{Feature} & \textbf{Cancer} & \textbf{Cancer$\rightarrow$BE} \\
    & \textbf{Extractor} & \textbf{PoI \%} & \textbf{PoI \%}\\
    \hline
    MICCAI 2015& A-KAZE & 30.34\% & 21.69\%\\\hline
	MICCAI 2015& SURF & 25.58\% & 23.05\%\\\hline
	MICCAI 2015& SIFT & 30.73\% & 23.04\%\\\hline
	Augsburg & A-KAZE & 53.77\% & 55.70\%\\\hline
	Augsburg & SURF & 42.97\% & 39.54\%\\\hline
	Augsburg & SIFT & 48.34\% & 44.06\%\\\hline
\end{tabular}
\end{center}
\label{t.cancer.pois.database}
\end{table}

\begin{figure*}[!htb]
\begin{center}
 \begin{tabular}{cc}
  \includegraphics[scale=0.1]{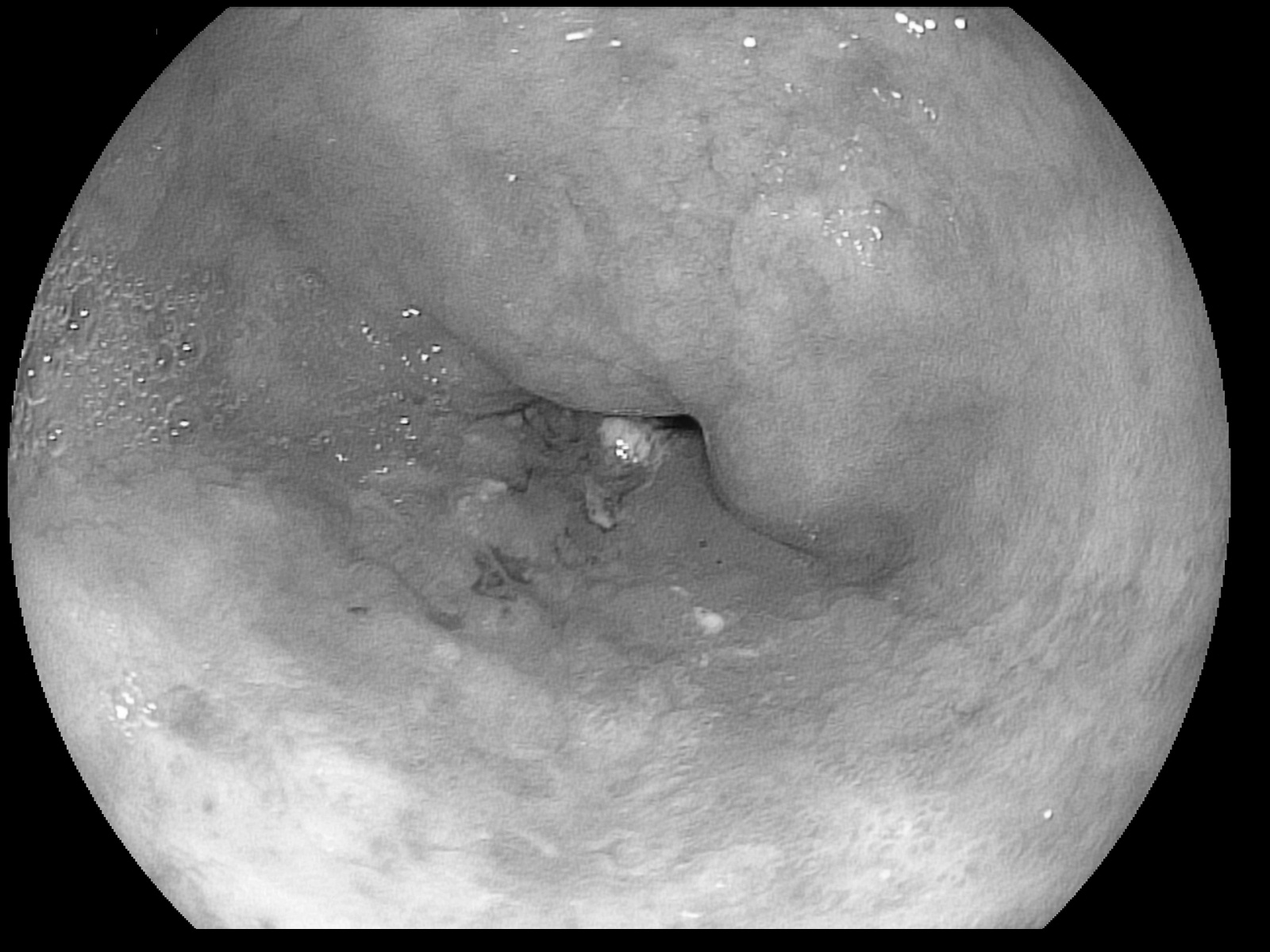} &
  \includegraphics[scale=0.1]{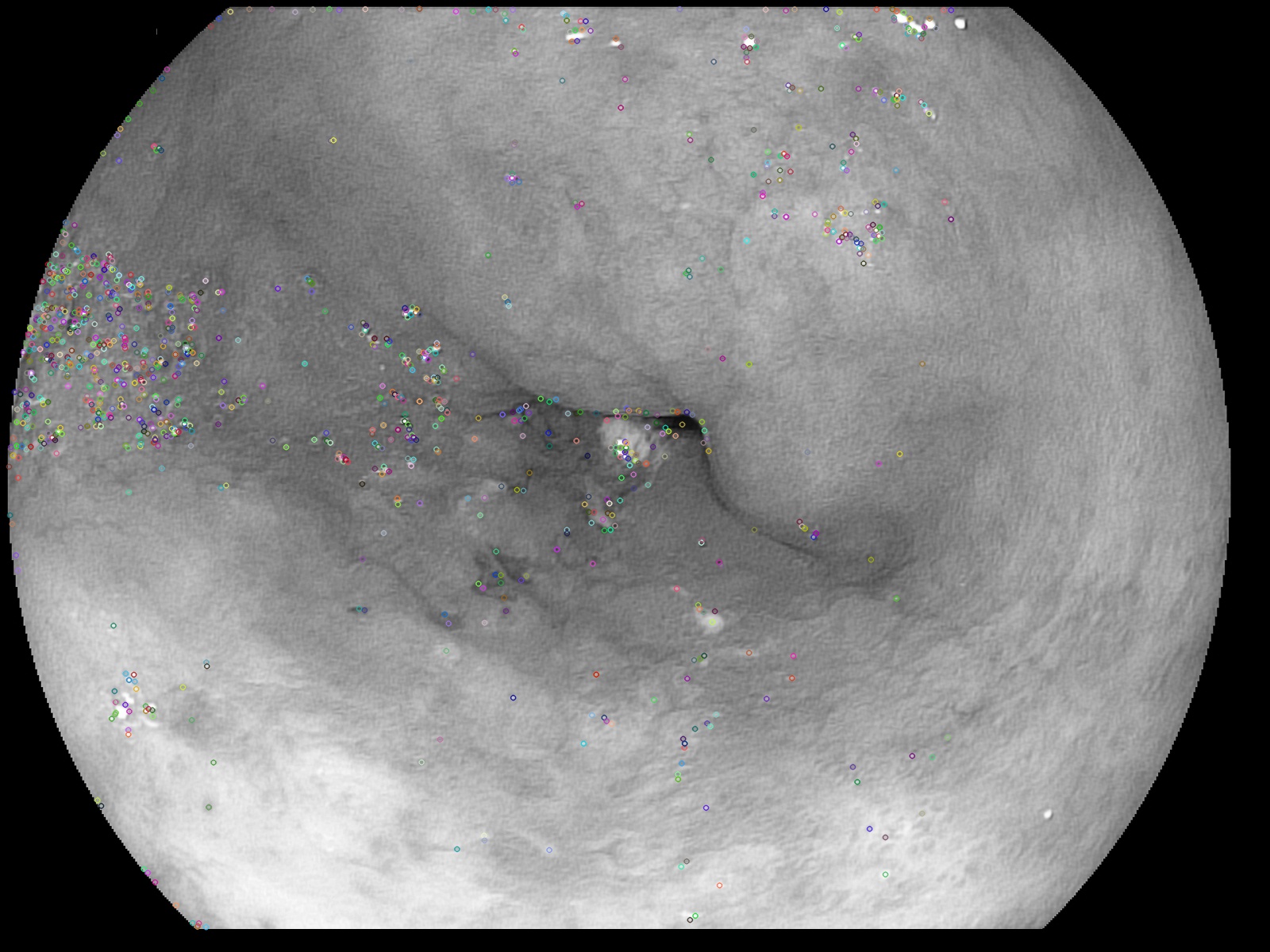}
  \\
  (a) & (b)
 \end{tabular}
 \begin{tabular}{c}	
 \includegraphics[scale=0.41]{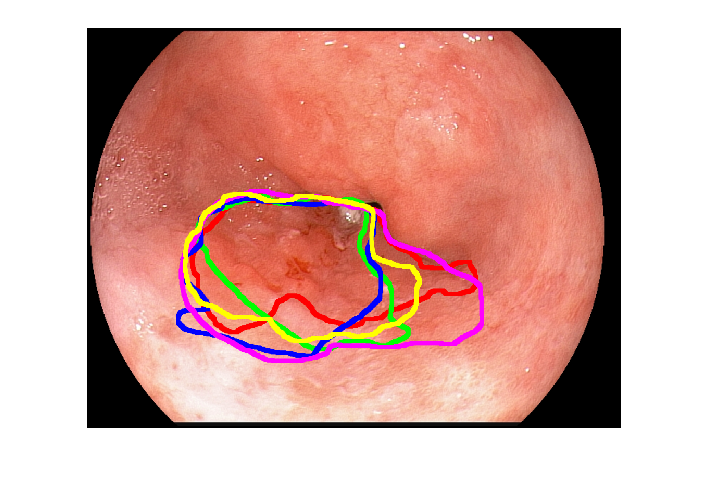}\\
 (c)
 \end{tabular}
\end{center}
 \caption{Misclassified image (patient 31) from MICCAI 2015 dataset: (a) gray-scale, (b) PoIs (SIFT), and (c) RGB version with delineations.} 
 \label{fig.misclass.miccai1}
\end{figure*}

\begin{figure*}[!htb]
\begin{center}
 \begin{tabular}{cc}
  \includegraphics[scale=0.11]{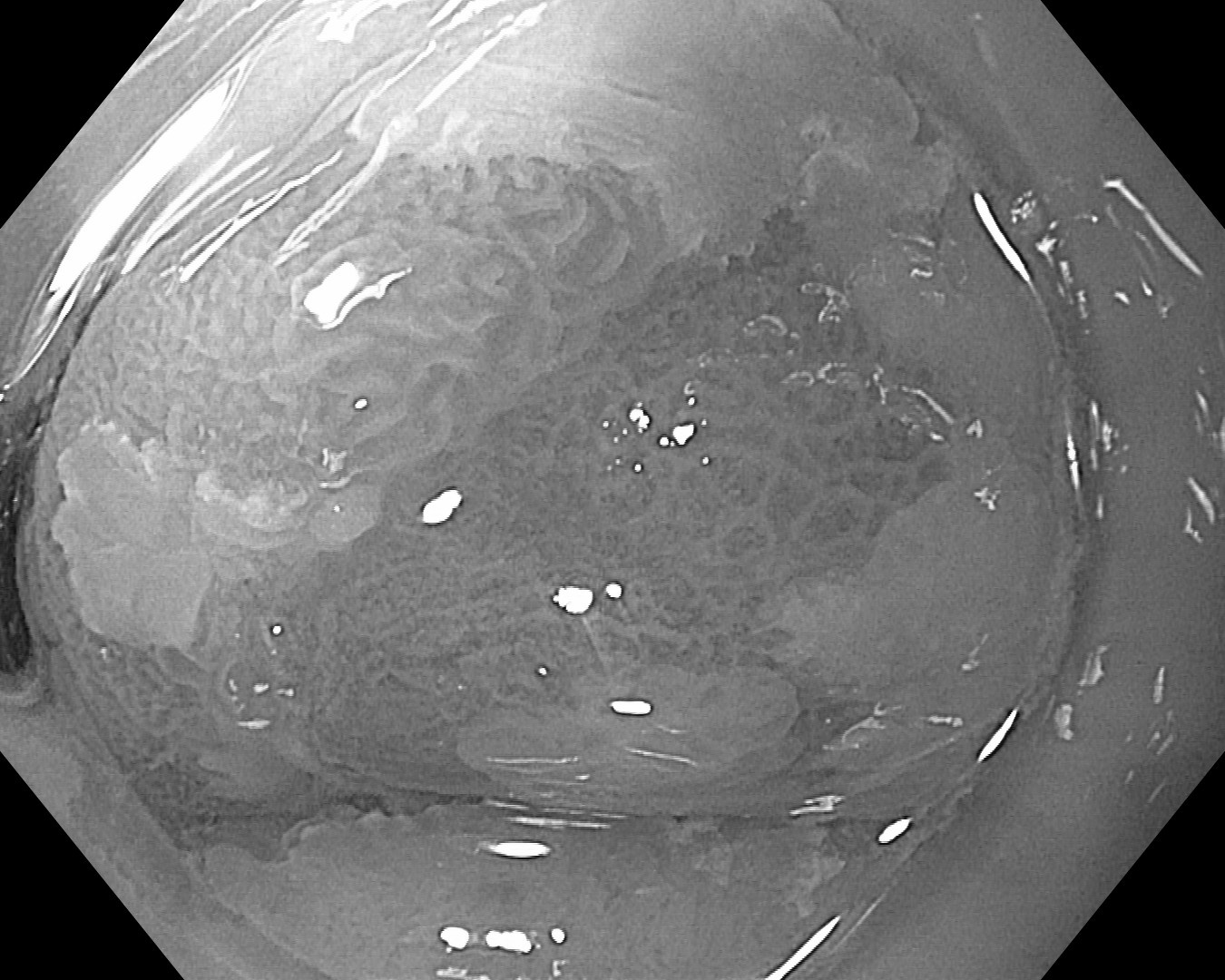} &
  \includegraphics[scale=0.11]{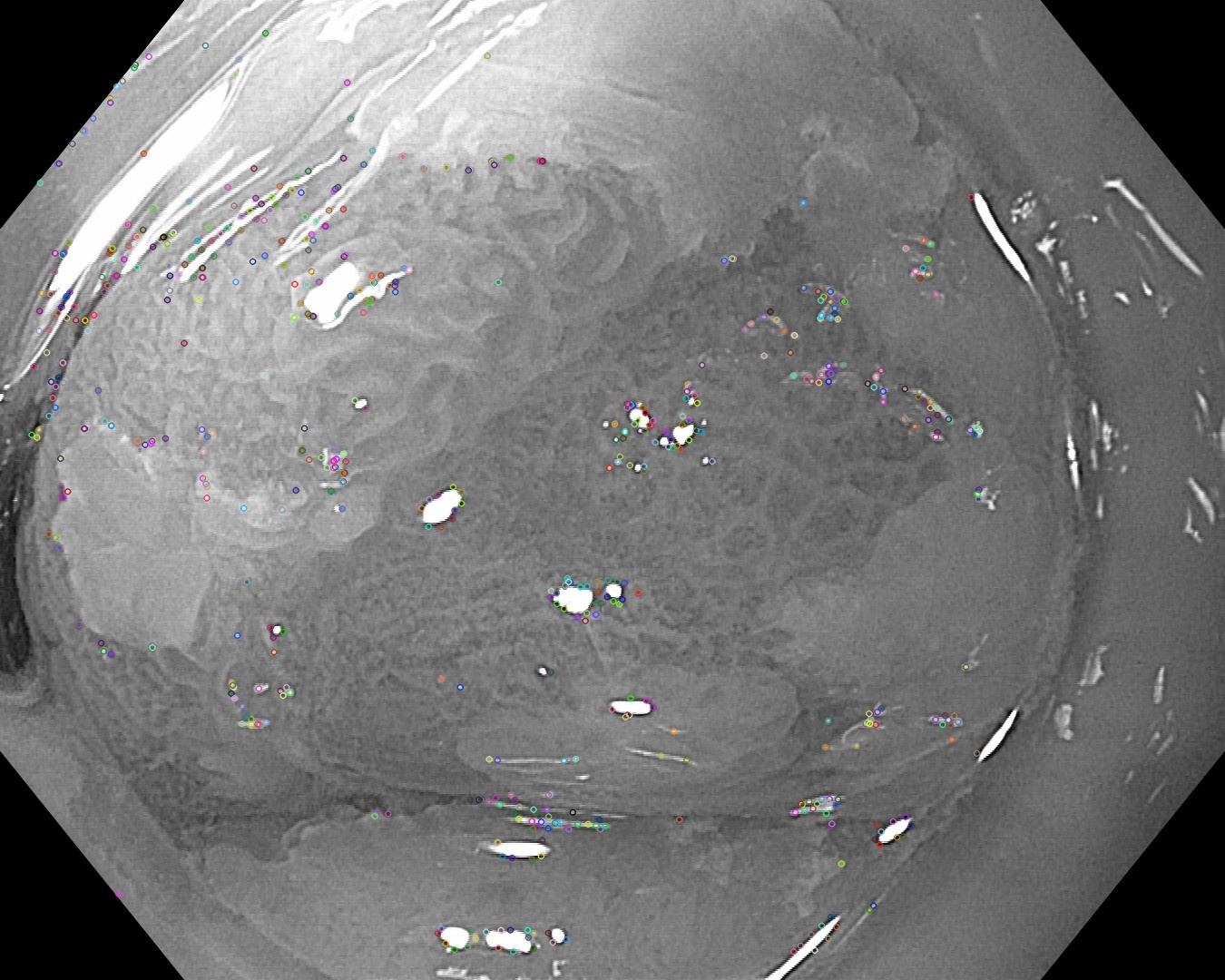}
  \\
  (a) & (b)
 \end{tabular}
 \begin{tabular}{c}	
 \includegraphics[scale=0.45]{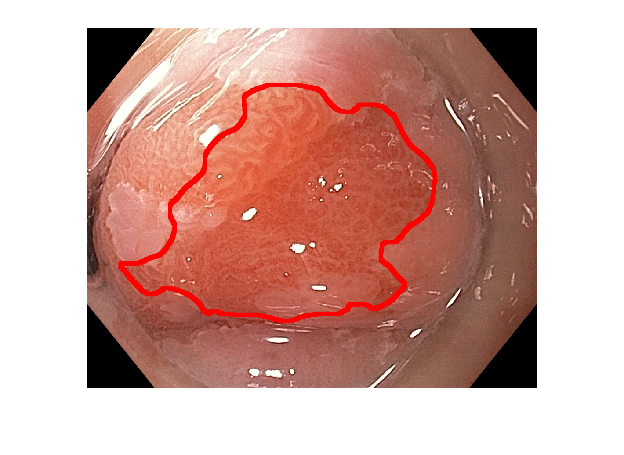}\\
 (c)
 \end{tabular}
\end{center}
 \caption{Misclassified image (patient 39) from Augsburg dataset: (a) gray-scale, (b) PoIs (A-KAZE), and (c) RGB version with delineation.} 
 \label{fig.misclass.augs2}
\end{figure*}

For visualization purposes, Figures~\ref{fig.misclass.miccai1} to~\ref{fig.misclass.augs2} depict some cancer patients that were misclassified as BE from both datasets. The PoIs showed in Figure~\ref{fig.misclass.miccai1} were calculated using SIFT and belong to the MICCAI 2015 dataset, and their percentage of incidence is $21.72\%$, which is slightly lower considering the average percentage presented in Table~\ref{t.cancer.pois.database} ($23,04\%$).

One can observe a considerable amount of PoIs located at the left-middle portion of Figure~\ref{fig.misclass.miccai1}b, mainly due to some air bubbles and foam. Problems with light (upper part of the image) also contribute to placing PoIs outside the delineated area. 

The PoIs showed in Figure~\ref{fig.misclass.augs2} were calculated using A-KAZE on an image from the Augsburg dataset. Their percentage of incidence is $7.5\%$, which is quite low considering the average percentage presented in Table~\ref{t.cancer.pois.database} ($53.77\%$). In this case, the main reason for placing PoIs outside the delineated area concerns illumination problems (brighter areas).

\section{Conclusions and Future Works}
\label{s.conclusions}

In this paper, we dealt with the problem of computer-assisted Barrett's esophagus identification \textcolor{black}{by means of bag-of-visual-words calculated using the OPF clustering technique. Such technique showed promising results, outperforming the previous handcrafted feature results in the same context. This suggests the generalization relevance of such technique, which can improve previous results in the same field not only for the BE context but for other in which the image representation configures the context to be evaluated.} BE stands for an illness that is likely to be confused with adenocarcinoma, and its early detection and prevention is of great concern.

\begin{sloppypar}We observed that only a very few works attempted at coping with the problem of automatic BE identification using computer vision and machine learning techniques to date. In this work, we fostered the research towards such area by introducing a supervised variant of the Optimum-Path Forest classifier for automatic BE recognition, as well as we showed how to build proper visual dictionaries using unsupervised OPF learning, \textcolor{black}{outperforming the results obtained in some recent works in which the same database and protocol were applied~\cite{Souza_SIBGRAPI:17,Souza:BVM2017}. Considering some previous works~\cite{Souza_SIBGRAPI:17,SouzaJr2018203,Souza:BVM2017}, the use of handcrafted features were based on the SURF and SIFT PoIs, but without the use of the OPF clustering as a way of dimensional reduction of the problem. Moreover, considering the improvements of the results, the use of the OPF clustering provides a new and promising way of BE and adenocarcinoma problem evaluation based on extracted key points}. \textcolor{black}{The presented results showed the relevance of such technique addressed to the BE and adenocarcinoma evaluation and description, contributing to the context literature and influencing the evaluation and description of other tissue diseases.} \textcolor{black}{Comparing the proposed method with others already published, we can ensure that with the use of the OPF for the BoVW step, improvements could be achieved considering the higher results obtained. Also, such technique provides advantages in the dimension reduction of the feature vector calculation, once even with a different number of key points per image, a standard method of feature calculation is established. Again, the OPF clustering may provide flexibility and time saving for such task.}\end{sloppypar}

The experimental results were considered over two datasets: (i) MICCAI 2015, and (ii) Augsburg. For both scenarios, we evaluated five classification techniques and three unsupervised learning approaches to build the visual dictionaries. Also, we considered dictionaries with three distinct sizes and even three different feature extractors. 

\begin{sloppypar}The experiments pointed out that bag-of-visual-words techniques are suitable to handle BE automatic identification, and there must be a trade-off between the number of visual words and the amount of information they can encode (i.e., size of the clusters). Additionally, both supervised and unsupervised OPF-based classifiers achieved the most accurate results, thus supporting the main contributions of this paper. \end{sloppypar}

\textcolor{black}{In the following, a bullet list of trends based on the achieved results is presented: 
\begin{itemize}
\item the OPF classifier presented the highest results of accuracy in all experiments, and may be highly recommended considering the high generalization that provided for such a context, even for different description scenarios;
\item the representation of BE and adenocarcinoma by means of image description techniques may provide encouraging results, and with less computation processing cost as needed in more sophisticated techniques; 
\item the A-KAZE features showed the very best results for BE and adenocarcinoma description in the Augsburg dataset evaluation, suggesting to be a very important technique for the description of such diseases;
\item the use of handcrafted features still has potential to evaluated for BE and adenocarcinoma problem, considering the several number of techniques, such as fisher vectors and sparse coding;
\item the use of OPF clustering improved the current results and can be applied to a large number of cases of image description of the BE and adenocarcinoma regions; 
\item the way of improving the selection of key points in each region (cancerous and non-cancerous) still shows potential considering the influence of the number of key points in each region for the correct classification result.
\end{itemize}
}

Regarding future works, we aim at considering deep learning and post-processing techniques after the construction of the bags, such as feature selection (i.e., visual word selection). \textcolor{black}{Additionally, this post-processing can be performed using the large number of machine learning techniques, such as SVM, OPF or even Convolution Neural Networks, providing intermediate learning for the dictionaries calculation. More techniques for image description are also considered to be evaluated using the bag-of-visual-words provided by the OPF clustering.}


\section*{Ackonowledgements}
\label{s.ack}
\begin{sloppypar}The authors are grateful to DFG grant PA 1595/3-1, Capes/Alexander von Humboldt Foundation grant number BEX 0581-16-0, CNPq grants 306166/2014-3 and 307066/2017-7, as well as FAPESP grants 2013/07375-0, 2014/12236-1, and 2016/19403-6. This material is based upon work supported in part by funds provided by Intel$^{\textregistered}$ AI Academy program under Fundunesp Grant No.2597.2017.\end{sloppypar}

\bibliographystyle{spmpsci}
\bibliography{paper}



\end{document}